\pdfoutput=1

\documentclass[11pt]{article}
\usepackage{arabtex}
\usepackage{url}

\usepackage[]{acl}

\usepackage{times}
\usepackage{latexsym}
\usepackage{arydshln}
\usepackage{utf8}
\setcode{utf8}
\usepackage[T1]{fontenc}

\usepackage[utf8]{inputenc}
\usepackage[T2A,LAE,T1]{fontenc}
\usepackage[arabic,USenglish]{babel}

\usepackage{microtype}

\usepackage{inconsolata}
\usepackage{appendix}
\usepackage{todonotes}
\usepackage{booktabs}

\usepackage{float} 
\usepackage{graphicx} 
\usepackage{caption} 
\usepackage{subcaption} 
\usepackage{rotating}
\usepackage{multirow}
\usepackage{tikz,comment}
\def\checkmark{\tikz\fill[scale=0.5](0,.35) -- (.25,0) -- (1,.7) -- (.25,.15) -- cycle;} 

 \usepackage{soul}

\usepackage{ragged2e}
\usepackage{array}

\title{{NADI} 2024: The Fifth Nuanced {A}rabic Dialect Identification Shared Task}

\author{Muhammad Abdul-Mageed,$^\lambda$$^\xi$ Amr Keleg,$^\sigma$ AbdelRahim Elmadany,$^\lambda$ Chiyu Zhang,$^\lambda$ \\ 
\textbf{Injy Hamed,$^\xi$$^\mu$ Walid Magdy,$^\sigma$ Houda Bouamor,$^\delta$  Nizar Habash$^\mu$$^\xi$}\\
$^\lambda$The University of British Columbia \quad 
$^\xi$MBZUAI\\
$^\sigma$The University of Edinburgh \quad 
$^\delta$Carnegie Mellon University in Qatar\\
$^\mu$New York University Abu Dhabi\\
  {\tt \{muhammad.mageed@,a.elmadany@,chiyuzh@mail.\}ubc.ca} \\
  {\tt \{a.keleg@sms,wmagdy@inf\}.ed.ac.uk}\\ 
  {\tt  injy.hamed@mbzuai.ac.ae ~ hbouamor@cmu.edu ~  nizar.habash@nyu.edu}\\
  }
\begin{document}
\maketitle
\begin{abstract}
We describe the findings of the fifth Nuanced Arabic Dialect Identification Shared Task (NADI 2024). NADI's objective is to help advance SoTA Arabic NLP by providing guidance, datasets, modeling opportunities, and standardized evaluation conditions that allow researchers to collaboratively compete on pre-specified tasks. NADI 2024 targeted both dialect identification cast as a multi-label task (Subtask~1), identification of the Arabic level of dialectness (Subtask~2), and dialect-to-MSA machine translation  (Subtask~3). A total of $51$ unique teams registered for the shared task, of whom $12$ teams have participated (with $76$ valid submissions during the test phase). Among these, three teams participated in Subtask~1, three in Subtask~2, and eight in Subtask~3. The winning teams achieved  50.57
F\textsubscript{1} on Subtask~1, 0.1403 RMSE for Subtask~2, and 20.44 BLEU in Subtask~3, respectively. Results show that Arabic dialect processing tasks such as dialect identification and machine translation remain challenging. We describe the methods employed by the participating teams and briefly offer an outlook for NADI.

\end{abstract}

\section{Introduction}\label{sec:intro}
\textit{Arabic} is a collection of languages, language varieties, and dialects that can be classified into three main categories. \textit{Classical Arabic} (CA) is the language of the Qur'an, old literature, and old scientific writing. CA has played a significant role in the spread of Islamic culture and continues to be crucial for scholarship in language and religious institutes. \textit{Modern Standard Arabic} (MSA) is a simplified and `modernized' descendent of CA that is employed in formal education and pan-Arab media as well as governmental circles in most Arab countries. \textit{Dialectal Arabic} (DA) refers to the various forms of Arabic spoken in different parts of the Arab world in informal settings, TV shows, and everyday life. These three main categories of Arabic share lexica and grammatical structures to varying degrees, with some dialects being at one end of the continuum and CA at the other end.  

The \textit{Nuanced Arabic Dialect Identification (NADI)} shared task series was launched in 2020 as a venue for creating resources, affording modeling opportunities, and building a research community around the processing of dialectal Arabic. NADI 2024 is the fifth version of the shared task, hosted by the Second Arabic Natural Language Processing Conference (ArabicNLP 2024).

\textit{Dialect identification} is the task of automatically detecting the source variety of a given text or speech segment. In previous versions of NADI, dialect identification was cast as a \textit{single-label} classification task. That is, the input text can be assigned only one dialectal category (usually at the country level). Arabic dialects, however, can overlap significantly. This is especially the case for dialects spoken in geographically proximate areas where lexica, grammatical structures, as well as sound patterns are usually shared to notable degrees. While speech language identification models, e.g.,~\cite{kulkarni-aldarmaki-2023-yet,sullivan2023robustness,radhakrishnan2023parameter}, would typically have access to acoustic features that can help facilitate teasing apart these neighboring varieties, this is not the case for text-based systems as the input is intrinsically impoverished. Aligning with this observation, recent work by~\newcite{keleg-etal-2023-aldi} analyzed the errors of a single-label dialect classification system and found that about 66\% of these errors are not true errors. To accommodate these research findings and open up a space for further investigation of challenges faced by single-label models, we design a subtask in NADI 2024 (Subtask~1) as a \textit{multi-label} classification task in which the given text can belong to more than a single Arabic dialect. Since Arabic dialects also overlap, sometimes significantly, with MSA, we also introduce a new subtask for estimating the level of \textit{dialectness} of text on a scale between zero and one (Subtask~2). Due to the challenges posed by dialects for machine translation (MT) systems, NADI 2024 continues to offer opportunities for advancing the translation of Arabic dialects through Subtask~3. We now review related literature.



\section{Literature Review}\label{sec:lit}
\subsection{Arabic Dialect Identification}
Unlike CA and MSA, which have been studied rather extensively~\cite{badawi1973mustawayat,Brustad:2000:syntax,Holes:2004:modern}, DA received attention relatively recently. Most early efforts focused on creating regional or country-level dialect datasets~\cite{diab2010colaba,Smaili:2014:building,Jarrar:2016:curras,Khalifa:2016:large,alsarsour-etal-2018-dart,el-haj-2020-habibi} and region-level dialect identification models from text~\cite{zaidan-callison-burch-2011-arabic,Elfardy:2014:aida,Meftouh:2015:machine,Bouamor:2018:madar,humayun2023transformer}.
Several works also introduced larger Twitter datasets covering dialects from 10-21 countries~\cite{Mubarak:2014:using,Abdul-Mageed:2018:you,Zaghouani:2018:araptweet, abdelali-etal-2021-qadi,issa-etal-2021-country,baimukan-etal-2022-hierarchical,Althobaiti_2022}, with some works such as~\citet{abdul-mageed-etal-2020-toward} introducing models targeting country, province, and city levels. Several benchmarks, e.g., ORCA~\cite{elmadany-etal-2023-orca} and DOLPHIN~\cite{nagoudi-etal-2023-dolphin}, involve several dialectal datasets. 

The NADI shared task continues to build on these previous efforts in offering datasets and affording modeling opportunities for identifying Arabic dialects~\cite{abdul-mageed-etal-2020-nadi,abdul-mageed-etal-2021-nadi,abdul-mageed-etal-2022-nadi,abdul-mageed-etal-2023-nadi}. This year, we employ a multi-label setting to take into account (i) suggestions by~\citet{keleg-magdy-2023-arabic} who highlight the limitations of addressing dialect identification as a single-label classification problem and propose defining it in a multi-label setting and (ii) issues of overlap in identical sentences across different dialects in the MADAR-26 test set~\cite{Bouamor:2018:madar} identified by~\citet{olsen-etal-2023-arabic}.


\subsection{Dialectness Level of Arabic}

DA does not exist in isolation from MSA. DI identification on the sentence level takes a binary view in distinguishing between MSA and DA. \citet{badawi1973mustawayat} identified five different levels of spoken Arabic in Egypt ranging from \textit{Heritage Classical Arabic} to \textit{Illiterate Colloquial Arabic}. He identified some phonological, morphological, lexical, and syntactic features of each of these levels. The same Arabic speaker employs different levels according to different sociolinguistic factors.


In an early work by \citet{zaidan-callison-burch-2011-arabic}, they asked crowdsourced annotators to identify the dialect and the level of dialectness of online comments to newspaper articles, forming the \textit{AOC} dataset. They have only provided four short descriptive labels for the levels of dialectness (\textit{No dialect, A bit of dialect, Mixed, Mostly Dialect}) and relied on the annotators' perceptions of these labels. Notably, their guidelines allow for assigning a high level of dialectness to a sentence having a highly dialectal word even if the remaining words are perceived to be less dialectal or in MSA, which is not the case for previous guidelines.

Recently, \citet{keleg-etal-2023-aldi} recycled the discrete level of dialectness labels from the \textit{AOC} dataset, transforming them into a continuous variable, Arabic Level of Dialectiness (\emph{ALDi}), taking values between 0 and 1, to form the \textit{AOC-ALDi} dataset. We decided to use the same operationalization of ALDi, while providing more elaborate guidelines to reduce the variability of the assigned ALDi levels for the same sentences.

\subsection{Arabic Machine Translation}
Several studies have addressed dialectal Arabic machine translation (MT), covering 
multiple dialects and translation directions \cite{abdul-mageed-etal-2023-nadi}. With regards to translation between Arabic variants, throughout the last year, numerous initiatives have focused on this task. 
Notably, the OSACT Dialect to MSA MT shared-task \cite{elneima-etal-2024-osact6} focused on translating Arabic dialects (Gulf, Egyptian, Levantine, Iraqi, and Maghrebi) into MSA. The proposed approaches mostly involved utilizing pretrained large language models (LLMs) 
with experimental designs incorporating zero-shot, few-shot, and fine-tuning setups, as well as data augmentation. In line with benchmarking LLMs, \citet{alam-etal-2024-codet} presented a benchmark with a focus on dialectal languages, including translating 25 Arabic dialects to MSA. 
Other researchers have focused on 
translation between specific Arabic dialects and MSA, 
including Egyptian \cite{faheem2024improving}, Tunisian \cite{kchaou2023hybrid}, and Algerian \cite{babaali2024breaking}. Despite the growing interest in this MT task, it received less attention compared to translation between Arabic and foreign languages, where several efforts involved evaluating and benchmarking LLMs \cite{kadaoui-etal-2023-tarjamat,nagoudi-etal-2023-dolphin,banimelhem2023chatgpt,abdelali-etal-2024-larabench,enis2024llm,alkhawaja2024unveiling}. 
To further advance research in MT across Arabic variants and build on our previous efforts, we include DA to MSA MT task again this year. 
\subsection{The History of NADI Shared Task}

\emph{NADI 2020}, the first NADI shared task~\cite{abdul-mageed-etal-2020-nadi} involved two subtasks, one targeting country level (21 countries) and another focusing on province level (100 provinces), both exploiting Twitter data. NADI 2020 was the first shared task to exploit naturally occurring fine-grained dialectal text at the sub-country level. 
\emph{NADI 2021}, the second version~\cite{abdul-mageed-etal-2021-nadi} targeted the same 21 Arab countries and 100 corresponding provinces as NADI 2020, also using Twitter data. However, it improved upon the previous version by removing non-Arabic data and distinguishing between MSA and DA. It introduced four subtasks: MSA-country, DA-country, MSA-province, and DA-province. 

\emph{NADI 2022}~\cite{abdul-mageed-etal-2022-nadi} continued the focus on studying Arabic dialects at the country level, but also included dialectal sentiment analysis with an objective to explore variation in socio-geographical regions that had not been extensively studied before.
Finally, \emph{NADI 2023}, the fourth edition~\cite{abdul-mageed-etal-2023-nadi}, proposed new MT subtasks from four dialectal Arabic varieties to MSA, in two themes: open-track and closed-track, as well as a dialect identification subtask at the country level.

In this paper, we 
introduce the
fifth edition of NADI by remodeling the dialect identification task into a multi-label classification task, introducing the ALDi estimation subtask, and continuing our MT subtask.
\section{Task Description}\label{sec:tasks}



\begin{table*}[!ht]
\setlength{\tabcolsep}{3pt}
 \resizebox{\textwidth}{!}{%
\tiny
\centering
\begin{tabular}{rcp{1.4cm}c}
\toprule

\multicolumn{1}{c}{\textbf{Sentence}} & \textbf{GEO} & \multicolumn{1}{c}{\textbf{Valid in}} & \textbf{ALDi} \\ 
\midrule
\RL{ اللهم انت ربي لا اله الا انت خلقتني وانا عبدك وانا علي عهدك ووعدك ما استطعت اعوذ بك من شر ما صنعت \#اذكار\_الصباح\_و\_المساء} & \textbf{SA} & \textbf{DZ}, \textbf{EG}, \textbf{JO}, \textbf{PS}, \textbf{SD}, \textbf{SY}, \textbf{TN}, \textbf{YE} & 0.00 \\


\RL{\#تريكه\_في\_كاس\_العالم شاهد ماذا قال \#الغندور علي المطالبه برجوع ابو تريكه للعب مع مصر في كاس العالم } & \textbf{EG} & \textbf{DZ}, \textbf{EG}, \textbf{JO}, \textbf{PS}, \textbf{SD}, \textbf{SY}, \textbf{TN}, \textbf{YE} & 0.15 \\

\RL{لمن الحياه ترسل ليك رساله }& \textbf{SD} & \textbf{PS}, \textbf{SD}, \textbf{YE} & 0.58 \\
\RL{وين يلعب هذا ما شفته }& \textbf{AE} & \textbf{DZ}, \textbf{PS}, \textbf{YE} & 0.64 \\
\RL{الحمد لله الجو برد الايمات يلي فاتوا الواحد مغموم اتقول مكسد بين فرادي تينه شمام } & \textbf{LY} & \textbf{DZ} & 0.83 \\
\RL{ايش دخل عارك ياحسن زميطه هذي العينات اللي يشتي له قيرعي } & \textbf{YE} & \textbf{YE} & 1.00 \\
\midrule
\end{tabular}%
 }
\caption{Sample sentences from MDID-DEV with their geolocated country (\textbf{GEO}), valid dialect labels (Subtask 1), and ALDi scores (Subtask 2). \textbf{DZ}: Algeria, \textbf{EG}: Egypt, \textbf{JO}: Jordan, \textbf{LY}: Libya, \textbf{PS}: Palestine, \textbf{SA}: Saudi Arabia, \textbf{SD}: Sudan, \textbf{SY}: Syria, \textbf{TN}: Tunisia, \textbf{AE}: UAE, \textbf{YE}: Yemen.}
\label{tab:examples}
\end{table*}

\begin{table*}[!ht]
 \renewcommand{\arraystretch}{1.3}
\resizebox{\textwidth}{!}{%
\centering
\begin{tabular}{lrr}
\toprule
\textbf{Dialect } & \multicolumn{1}{r}{\textbf{Source (Dialect)~~~~~~~~~~~~~~~~~~~~~~~~~~~~~~~~~~~~~~~}} & \multicolumn{1}{r}{\textbf{Target (MSA)~~~~~~~~~~~~~~~~~~~~~~~~~~~~~~~~~~~~~~~}} \\
\bottomrule
\multicolumn{1}{c}{\multirow{3}{*}{\rotatebox[origin=c]{90}{Egyptian}} } & \RL{ أنا بس عايزك ترتبي نفسك من دلوقتي إنك تبقي زي ستات عيلة الدالي. } & \RL{ أنا أريدك أن ترتبي نفسك من الآن أن تكوني مثل سيدات عائلة الدالي.} \\
\multicolumn{1}{c}{}  & \RL{ طب ايه بقى اللي انت قولته و خليت الريس يقلب عليك بالشكل ده؟} & \RL{حسنا، ما هذا الذي قلته، وجعلت الرئيس ينقلب عليك بهذا الشكل؟} \\
\multicolumn{1}{c}{}  & \RL{بلا مسقعة بلا رز بلبن اقعد ساكت بلاش كلام فارغ. } & \RL{لا مسقعة، ولا أرز بلبن. اسكت، وبلا كلام فارغ.} \\

\midrule

\multicolumn{1}{c}{\multirow{3}{*}{\rotatebox[origin=c]{90}{Emirati}} } & \RL{ أنا يا بنيتيه، حيث إنه ريولي عورتني من كثر اليلسة}& \RL{ أنا يا ابنتي، رجلي تؤلمني من الجلوس لمدة طويلة} \\
\multicolumn{1}{c}{}  & \RL{خله يولي، هذا كله مغربلنا و لا مبهدلنا، ما يشبع، و بعدين ليش خايفين ؟  } & \RL{دعه وشأنه، لقد ضرنا وعذبنا، لا يشبع، ومما أنتم خائفون؟ } \\
\multicolumn{1}{c}{}  & \RL{ يعني ألحين إنتي ما تعرفين شي عن سالفة الرضاعة ؟} & \RL{معنى ذلك أنك لا تعلمين شيئا عن قصة الرضاعة؟} \\

\midrule

\multicolumn{1}{c}{\multirow{3}{*}{\rotatebox[origin=c]{90}{Jordanian}} } & \RL{ طمعانين! الشاطر ينهب، و هو راخي لهم الحبل و ساكت!}& \RL{ إنهم طماعون! الكل يسرق، وهو متهاون ولا يتكلم.} \\
\multicolumn{1}{c}{}  & \RL{مبدئيًا رح أكتب لك هاظ العلاج مشان يخفف عليها، و رح نشوف شو بصير معانا } & \RL{في البداية سأصف لك هذا الدواء لتخفيف الآلام، وسنرى النتيجة.} \\
\multicolumn{1}{c}{}  & \RL{ ما بدنا اياك تتورط، و افهم من الدكتور ايش اللي بده اياه منك} & \RL{لا نريد أن تتورط، وافهم من الطبيب ما الذي يريده منك.} \\

\midrule

\multicolumn{1}{c}{\multirow{3}{*}{\rotatebox[origin=c]{90}{Palestinian}} } & \RL{ يم مجيتش تمرظ إلا هسا؟ ترا انته لا للسدي و لا للهدي، مثلك مثل مسعود الحذر}& \RL{ يعني لم تمرض إلا الآن؟ بالمناسبة أنت عديم الجدوى، مثلك مثل مسعود الحذر.} \\
\multicolumn{1}{c}{}  & \RL{ بس انا بدي اياك تروحي عند ابو العبد تخليه يشوف ابو يزن يتنازل عن حقو} & \RL{لكن أنا أريدك أن تذهبي عند أبي العبد وتجعليه يتنازل عن حقه.} \\
\multicolumn{1}{c}{}  & \RL{ فش عندهم شرش هالذكاء، من وين جايبيته اه قوليلي؟ اكيد مني يعني طالعة علي!} & \RL{ليس عندهم ذكاء، من أين أحضرته؟ أخبريني. من المؤكد أنه مني، يعني أنت مثلي!} \\

 \bottomrule
\end{tabular}%
}
\caption{Random examples from MT-2024-DEV dataset spanning the four covered dialects.}
\label{tab:examples_mt}
\end{table*}


NADI 2024 maintains the focus on processing dialectal Arabic. More concretely, we target both dialect identification (DID) and dialectal machine translation through three subtasks. \textbf{Subtask~1} focuses on DID, cast as a multi-label classification task, and \textbf{Subtask~2} aims at capturing the Arabic level of dialectness in texts (ALDi). As translation of Arabic dialects remains particularly challenging, we devote \textbf{Subtask~3} to dialect MT. We now describe each subtask in detail.

\subsection{Subtask~1 --  Multi-Label Dialect ID}\label{sec:MLDID_task}

In this subtask, we propose multi-label dialect identification (MDID) at the country level. The objective is to evaluate the feasibility of using single-label Arabic dialect identification datasets to train a multi-label system that can predict \textit{all} dialects in which a given sentence is valid.

\paragraph{Tranining Data} We provide participants with the \textit{training} splits of following datasets: \texttt{MADAR-2018}~\cite{bouamor-etal-2019-madar} \texttt{NADI-2020-TWT}, \texttt{NADI-2021-TWT}, and \texttt{NADI-2023-TWT}~\cite{abdul-mageed-etal-2020-nadi, abdul-mageed-etal-2021-nadi, abdul-mageed-etal-2023-nadi}. 

\paragraph{Dev and Test Data} 
We provide a new multi-label development set: \texttt{MDID-DEV}, henceforth \texttt{MDID-DEV} for brevity, as explained in §\ref{sec:datasets}. This dataset has $120$ samples with manually assigned validity labels of eight different Arab countries: \textit{Algeria, Egypt, Jordan, Palestine, Sudan, Syria, Tunisia,} and \textit{Yemen}. Examples of those sentences are provided in Table \ref{tab:examples}. We do not restrict systems to these eight dialects. Hence we include two undisclosed dialects in our test data and ask participants to develop their models such that they can predict all valid dialects out of the $18$ country-level ones from NADI 2023. 
The undisclosed dialects are \textit{Iraq} and \textit{Morocco}. Accordingly, the \texttt{MDID-TEST} set contains 1,000 samples covering nine dialects.\footnote{We also note that one of the Jordanian annotators did not complete the labeling process on time, and so we did not include the labels from Jordanian annotators in the test sets.}

\paragraph{Restrictions} Subtask~1 operates under a \textbf{\textit{closed-track}} policy where participants are allowed to \textit{only} use the datasets we provide for system training.

\subsection{Subtask~2 --  ALDi Estimation}
\newcite{keleg-etal-2023-aldi} define the Level of Dialectness as the extent by which a sentence diverges from the Standard Language. We use their operationalization to estimate the ALDi of sentences as a continuous value in the range [0, 1]; where 0 means MSA and 1 implies high divergence from MSA.

\paragraph{Training Data} We provided the teams with AOC-ALDi dataset's training split \cite{keleg-etal-2023-aldi}. 

\paragraph{Dev and Test Data} 
The dev and test sets collected for Subtask~1 were extended with a second layer of annotation for manual ALDi levels, forming \texttt{ALDi-DEV} and \texttt{ALDi-TEST} sets. The annotation process is outlined in §\ref{sec:datasets}.
\paragraph{Restrictions} Subtask~2 operates under an \textbf{\textit{open-track}} policy, allowing participants to train their systems on any additional datasets of their choice, provided that they explain the sources of the data and how it is used and that these additional training datasets are public at the time of submission.

\subsection{Subtask~3 -- Machine Translation} 

Similar to NADI 2023, this subtask is focused on machine translation from four Arabic dialects (i.e., \textit{Egyptian}, \textit{Emirati}, \textit{Jordanian}, and \textit{Palestinian}) to MSA at the sentence level. Unlike NADI 2023 where we had a close-track version of the MT task, we exclusively offer an open-track theme this year.

\paragraph{Training Data} We do not provide direct training data. However, to facilitate Subtask~3, we point participants to the MADAR parallel dataset\footnote{MADAR dataset can be acquired directly 
at \href{https://camel.abudhabi.nyu.edu/madar-parallel-corpus/}{MADAR Parallel Corpus}. It comprises parallel sentences encompassing the dialects of 25 cities from the Arab world, as well as English, French, and MSA. Participants are permitted to use only the Train split of the MADAR parallel data for this subtask and must report on the Dev and Test sets we provide. The use of MADAR Dev and Test sets is not allowed in this subtask.}~\cite{bouamor-etal-2019-madar} for system training and a monolingual dataset\footnote{The monolingual dataset is composed of the training splits of NADI 2020, NADI 2021, and NADI 2023, comprising 20k, 20k, and 18K tweets, repectively.} that participants can manually translate and use for training. 


\paragraph{Dev and Test Data} 
For Subtask~3, we manually curated and translated completely new development and test data that were not used in NADI-2023. The development split, MT-2024-Dev, comprises $400$ sentences, with $100$ sentences representing each of the four dialects, while test split, MT-2024-Test, totals $2,000$ sentences, with $500$ from each dialect. Table~\ref{tab:examples_mt} shows example sentences from MT-2024-Dev for each of the four countries. During the competition, we intentionally kept the source domain of these datasets undisclosed.\footnote{Since we typically maintain a live leaderboard for post-competition evaluation, we not disclose the \texttt{MT-2024} data domain here either.}

\paragraph{Restrictions} Subtask~3 operates under an \textbf{\textit{open-track}} policy, allowing participants to train their systems on any additional datasets of their choice, provided these additional training datasets are public at the time of submission. For example, participants were permitted to manually create new parallel datasets. For transparency and the benefit of the wider community, we required participants to submit the datasets they created along with their test set submissions.

\subsection{Evaluation Metrics}\label{subsec:dataset-eval-metrics}

The official evaluation metric for Subtask~1 is the macro-averaged \texttt{F\textsubscript{1}} score. More specifically, we compute the \texttt{F\textsubscript{1}} score independently for each country in the evaluation dataset (eight for the development set and nine for the test set), then compute the average of these individual-country \texttt{F\textsubscript{1}} scores.\footnote{Pariticiating teams submitted validity predictions for the 18 countries of the training sets. We plan to rerun the evaluation upon collecting labels for more country-level dialects.} Additionally, we report system performance in terms of \texttt{Precision}, \texttt{Recall}, and \texttt{Accuracy} for submissions to Subtask~1. The metric for Subtask~2 is the \texttt{Root Mean Square Error (RMSE)}. For Subtask~3, we use the BLEU score~\cite{papineni2002bleu} as the official metric.\footnote{We plan to consider other MT evaluation metrics in future versions of the shared task.} We calculate the overall BLEU score over all the samples (i.e., across all dialects) to rank the submitted systems for Subtask~3. We also present BLEU scores calculated separately for each of the four dialects (i.e., \textit{Egyptian}, \textit{Emirati}, \textit{Jordanian}, and \textit{Palestinian}). 

\subsection{Submission Rules}\label{subsec:dataset-eval-roles}
We allowed participant teams to submit up to \textit{five} runs for each test set, for each of the three subtasks. For each team, only the submission with the highest score was retained. While the official results were exclusively based on a blind test set, we requested participants to include their results on Dev splits in their papers. To facilitate the evaluation of participant systems, we established a CodaLab competition for scoring each subtask (i.e., a total of three CodaLabs).\footnote{Our CodaLabs are available at the following links: \href{https://codalab.lisn.upsaclay.fr/competitions/18130}{\texttt{Subtask~1}}, \href{https://codalab.lisn.upsaclay.fr/competitions/18135}{\texttt{Subtask~2}}, and \href{https://codalab.lisn.upsaclay.fr/competitions/18133}{\texttt{Subtask~3}}.} Similar to previous NADI editions, we are keeping the CodaLab for each subtask active even after the official competition has concluded. This is to encourage researchers interested in training models and assessing systems using the shared task's blind test sets.
A nuance is that since subtasks~1 and~2 are new to NADI with limited training data available publicly, we 
 share the individual labels of the development/test sets for these two subtasks.\footnote{We will be glad to consider collaborations on extending the datasets for all our subtasks to other country-level dialects.}

\section{Evaluation Data for Subtasks 1 and 2}\label{sec:datasets}

\subsection{Samples Curation}\label{sec:sampling_sentences}
We employ the same methods as in~\newcite{abdul-mageed-etal-2023-nadi} to collect geolocated tweets, then randomly sample 80 data points for the following ten countries from which we could recruit annotators:\footnote{As per §\ref{sec:MLDID_task}, labels from Jordan are only in the Dev set.} \textit{Algeria, Egypt, Iraq, Jordan, Morocco, Palestine, Sudan, Syria, Tunisia,} and \textit{Yemen}, in addition to 80 data points from four other Arab countries \textit{Lebanon, Libya, Saudi Arabia, UAE}. 
These additional samples 
are expected to be labeled as invalid in the dialects of the ten countries from which we recruited the annotators. Including them ensures the dataset's samples cover a wider range of dialects.

We use an in-house MSA/DA classification model (acc=89.1\%, \texttt{$F_1$} score=88.6) introduced in~\newcite{abdul-mageed-etal-2021-arbert} to ensure that for each country's 80 geolocated samples, five are in MSA, and 75 are in DA. The overall dataset size for the shared task is 1,120 samples. Each annotator labeled the whole dataset. 
We remove user mentions, URLs, and emojis from the data, but retain the hashtags, before labeling the samples. We annotate our dataset on Upwork, incurring a total cost of $\$1,700$. 

\subsection{Annotation Process}\label{sec:annotation_process}
For Subtask~1, we follow \citet{keleg-magdy-2023-arabic}'s proposal for building multi-label ADI datasets, mainly by asking native speakers of different Arabic dialects (on the country level) to check if each sentence is valid in one of the dialects spoken in their countries or not. For Subtask~2, we decided to provide more elaborate definitions for the different levels of dialectness than those in the guidelines of~\citet{zaidan-callison-burch-2011-arabic}.\footnote{Refer to §\ref{sec:annotation_guidelines} of the Appendix for further details.}
%
%
For each tweet, we ask two questions:\\
\textbf{Q1)} \textit{Is it possible that the tweet is authored by someone who speaks one of your country's dialects?}

\textbf{{Options:}} (a) \textit{Yes}, (b) \textit{Not Sure/Maybe}, or (c) \textit{No}.\\
\textbf{Q2)} \textit{What is the Arabic Level of Dialectness (ALDi) of the tweet?} We define the following levels:
\vspace*{-0.3cm}
\begin{enumerate}
    \setcounter{enumi}{-1}
    \setlength{\parskip}{0pt}
    \setlength{\itemsep}{0pt}
    \item \textbf{Sound MSA}: Tweets written in fluent MSA.
    \item \textbf{Formal Colloquial or Colloquial-influenced MSA}: Tweets written in a language close to MSA but using some colloquial expressions (lexemes/ morphemes).
    \item \textbf{Natural/Ordinary Colloquial}: Tweets written in a colloquial language that is accepted and understood by all members of society of all ages and social/educational levels.
    \item \textbf{Informal (or Vulgar) Colloquial}: Tweets written in a colloquial language having expressions that are not accepted or understood by all members of society. It does not have to be vulgar or weak.
\end{enumerate}

We believe an Arabic speaker identifying the ALDi of a tweet needs to be familiar with the dialect in which the tweet is written. For this reason, an annotator is allowed to identify ALDi only if their answer to the first question (validity of tweet in one of their country-level dialects) is either \textit{Yes} or \textit{Not Sure/Maybe}.\footnote{See §\ref{sec:annotation_process} for information about the annotation process.}

For each of the ten specified countries, we managed to recruit three native speakers through Upwork to label all the 1,120 tweets.
Before inviting the annotators to the main task, we ask them to complete an onboarding task to get them acquainted with our objectives and clarify any misunderstandings. Afterward, the main task annotation process is split into five batches, 224 samples each, where feedback is provided to the annotators after each batch to ensure high quality. 
Annotators were paid \$8 after successfully completing each of the six tasks in addition to a bonus value between \$8 and \$12 after completing the whole process. After accounting for the platform fees, annotating the dataset cost about \$1,700.

\begin{table}[tb]
    \centering
    \tabcolsep5pt
    \resizebox{0.89\columnwidth}{!}{%
    \begin{tabular}{lcccc}
    \toprule
    \multirow{2}{*}{\textbf{Country}} & \multicolumn{3}{c}{\textbf{Subtask 1}} & \textbf{Subtask 2}\\ \cmidrule{2-5}
    & \textbf{Fleiss} $\mathbf{\kappa}$ & \textbf{N valid} & \textbf{N ¬valid} & \textbf{Krip.} $\mathbf{\alpha}$ \\
    \midrule
        Algeria &     0.56 & 333 (205) &  787 (666) &     0.66 \\
        Morocco &     0.62 & 230 (152) &  890 (784) &     0.74 \\
        Tunisia &     0.67 & 189 (129) &  931 (879) &     0.75 \\
        Egypt &     0.69 & 353 (257) &  767 (682) &     0.82 \\
        Sudan &     0.67 & 393 (283) &  727 (619) &     0.66 \\
        Palestine &     0.59 & 375 (245) &  745 (587) &     0.68 \\
        Syria &     0.54 & 475 (305) &  645 (543) &     0.79 \\
        Iraq &     0.61 & 271 (171) &  849 (738) &     0.73 \\
        Yemen &     0.52 & 454 (291) &  666 (477) &      0.50 \\
    \bottomrule
    \end{tabular}
    }
    \caption{Interannotator agreement scores -- Fleiss' Kappa ($\kappa$) for Subtask~1 and Krippendorff's Alpha-interval method ($\alpha$) for Subtask~2 -- for the full dataset. We also report the number of valid, not valid sentences out of the 1,120 according to majority voting, while showing the number of sentences with complete agreement (between brackets). \textbf{Note:} The country-level Krippendorff's Alpha scores are computed for their respective country's valid samples.}
    \label{tab:IAA_agg}
\end{table}

\paragraph{Interannotator Agreement (IAA) Scores} We use Fleiss' Kappa ($\kappa$) \cite{fleiss1971measuring}, and Kripendorff's Alpha ($\alpha$) \cite{krippendorff2004content} for measuring the IAA. The country-level scores in Table~\ref{tab:IAA_agg} indicate moderate to substantial agreement between the annotators for both subtasks. Moreover, there is not a noticeable variation among the scores across the countries, except for the $\alpha$ score for the Yemen annotators which is slightly lower than those of the other countries. We noticed the the Yemeni annotators had different perceptions of what counts as \textit{Sound MSA (Level 0)} and what counts as \textit{Natural/Ordinary Colloquial (Level 2)}.

\subsection{Label Aggregation Techniques}\label{sec:label_agg}
\paragraph{Subtask~1} A sentence is considered valid in a country-level dialect if among the three annotators from the respective countries: a) one of them answered \textit{Yes}, and b) another answered \textit{Yes} or \textit{Maybe}. On average, the same-country annotators fully agreed on the validity of more than 66\% of the valid samples, and the invalidity of more than 85\% of the invalid samples, as per Table~\ref{tab:IAA_agg}.

\paragraph{Subtask~2}
For each sentence, the ordinal ALDi levels assigned by the annotators from the different countries are aggregated into a single numeric value $\in [0, 1]$. Discrete ALDi levels (0, 1, 2, 3) are transformed into the following numeric values ($0, \frac{1}{3}, \frac{2}{3}, 1)$. The mean of these numeric values is used as the overall ALDi score for the sentence.

As mentioned in §\ref{sec:annotation_process}, annotators only assigned ALDi levels to sentences they rated as valid in their country-level dialect. Consequently, the number of ALDi annotations per sentence can range from 0 to 3*N where N is the number of countries from which annotators are recruited.
If a sentence is deemed invalid according to the majority vote label ( Subtask~1) for a country-level dialect, we discard the resective ALDi annotations (if any) assigned by the annotators' of this country.

\subsection{Formation of Development/Test Sets}\label{sec:split_analysis}
We used 120 samples from the first batch as the development sets (\texttt{MDID-DEV}, \texttt{ALDi-DEV}) shared with the participating teams. The first batch's remaining samples and the samples of the 4 succeeding batches form the test sets (\texttt{MDID-TEST}, \texttt{ALDi-TEST}).
For \texttt{ALDi-DEV} and \texttt{ALDi-TEST}, samples that are not valid in the considered dialects of the corresponding set have no assigned ALDi scores, and thus are not released as part of the dataset.

\paragraph{Analysis of the Development Sets}

\begin{figure}[tb]
    \centering
        \includegraphics[width=0.85 \columnwidth]{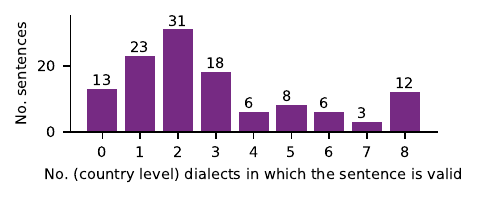}
        \caption{Distribution of the number of valid (country level) dialects out of 8 countries.}
        \label{fig:subtask1_distrubution}
\end{figure}


\begin{figure}[tb]
    \centering
    \includegraphics[width=0.9 \columnwidth]{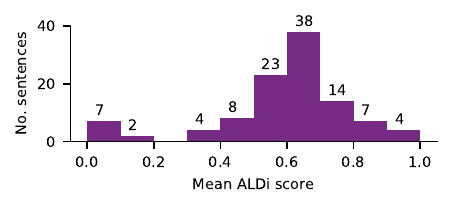}
    \caption{\texttt{ALDi-DEV}'s scores.}
    \label{fig:subtask2_distrubution}
\end{figure}

Figure~\ref{fig:subtask1_distrubution} shows that 13 samples are labeled as invalid in all of the 8 considered countries in \texttt{MDID-DEV}, and 23 samples are valid in only one dialect. We also report that 84 samples are valid in 2 or more country-level dialects (i.e., 70\% of the samples). This percentage is only expected to increase when validity annotations from other country-level Arabic dialects are collected. 

The aggregated ALDi scores have a multimodal distribution as per Figure~\ref{fig:subtask2_distrubution}. The first mode is related to the automatically identified MSA samples in the development set (8 in total). All of these samples are assigned ALDi scores \textless 0.2, and are judged as valid in all the considered country-level dialects. Conversely, the ALDi scores for the automatically identified DA samples are distributed around a score of 0.66 (the numeric value corresponding to Level 2 (\textit{Natural/Ordinary Colloquial}).


\section{Shared Task Teams \& Results}\label{sec:results}

\begin{table}[t]
\centering
\resizebox{\columnwidth}{!}{%
\setlength{\tabcolsep}{2pt}

\begin{tabular}{@{}llc@{}}
\toprule
\multicolumn{1}{c}{\textbf{Team}} & \multicolumn{1}{c}{\textbf{Affiliation}}   & \textbf{Tasks} \\ \midrule
AlexUNLP-STM~\cite{alexunlp-2024-nadi}                      & Alexandria University, Egypt               & 2              \\
Alson~\cite{alson-2024-nadi}                             & -, KSA                                     & 3              \\
Arabic Train~\cite{arabictrain-2024-nadi}                      & MBZUAI, UAE                                & 3              \\
ASOS~\cite{asos-2024-nadi}                              & Prince Sultan University, KSA              & 2, 3           \\
CUFE~\cite{cufe-2024-nadi}                              & Cairo University, Egypt                    & 2, 3           \\
dzNLP~\cite{dznlp-2024-nadi}                             & USTHB, Algeria                             & 1              \\
Elyadata~\cite{elyadata-2024-nadi}                          & Elyadata, Tunisia                          & 1              \\
MBZUAI BADG                       & MBZUAI, UAE                                & 3              \\
MBZUAI BLEU                       & MBZUAI, UAE                                & 3              \\
NLP\_DI~\cite{NLPDI-2024-nadi}                            & Dalle Molle Ins. of A.I.,   & 1              \\
                                                          &  Switzerland &              \\
Shaheen                           & MBZUAI, UAE                                & 3              \\
VBNN                              & MBZUAI, UAE                                & 3              \\ \bottomrule
\end{tabular}%
}
\caption{List of teams that participated in NADI-2024 shared task. Teams with accepted papers are cited.}
\label{tab:participating_teams}
\end{table}
\subsection{Participating Teams}\label{subsec:teams}
We received a total of 51 unique team registrations. At the testing phase,  a total of 76 valid entries were submitted by 12 unique teams. The breakdown across the subtasks is as follows:
\textit{ten} submissions for Subtask~1 from \textit{three} teams, 
\textit{seven} submissions for Subtask~2 from \textit{three} teams,
and 16 submissions for Subtask~3 from \textit{eight} teams. 
%
Table~\ref{tab:participating_teams} lists the 12 teams. 
We received eight description papers, all of which were accepted for publication.

\subsection{Baselines}\label{subsec:baselines}
We developed baseline (BL) models for each subtask for comparison against the teams' systems, as described below. These models were not shared with participating teams during the competition.

\paragraph{Subtask~1 Baselines} We use the softmax of a fine-tuned single-label DI system's logits to develop two baselines.\footnote{The fine-tuned baseline model can be accessed through \url{huggingface.co/AMR-KELEG/NADI2024-baseline}.} The first predicts the most probable labels such that their cumulative probability is $>90\%$. The second assumes the sentence is only valid in the most probable prediction. Lastly, we implement a Random baseline that generates random binary predictions for the validity of the sentences in the considered dialects.

\paragraph{Subtask~2 Baselines} We first use the Sentence ALDi model developed by \citet{keleg-etal-2023-aldi} as our supervised baseline. The second baseline is based on the distribution of the ALDi scores for the development set (Figure~\ref{fig:subtask2_distrubution}), where we implement a model that generates a constant score of 0.67 for all the sentences. In the third baseline, we use a Random ALDi generator ($\in [0, 1]$).

\paragraph{Subtask~3 Baselines} 

We extract parallel dialectal-to-MSA data of four dialects from MADAR-18 for training MT baselines for Subtask~3. We then fine-tune three baselines on the extracted data. These are \textit{AraT5\textsubscript{v2}}~\cite{elmadany-etal-2023-octopus, nagoudi-etal-2022-arat5}, \textit{mT5}~\cite{xue-etal-2021-mt5}, and \textit{AraBART}~\cite{kamal-eddine-etal-2022-arabart}.

\begin{table}[tb]
    \centering
    \setlength{\tabcolsep}{2pt}
\resizebox{\columnwidth}{!}{%
    \begin{tabular}{clcccc}
    \toprule
        \multirow{2}{*}{\textbf{Rank}}& \multirow{2}{*}{\textbf{System}} & \multicolumn{4}{c}{\textbf{Macro-average}}\\ \cmidrule{3-6}
        & & \textbf{Accuracy ($\uparrow$)}& \textbf{Precision ($\uparrow$)}& \textbf{Recall ($\uparrow$)}& \textbf{\texttt{F\textsubscript{1}} score ($\uparrow$)}\\
        \midrule
            1 & \multirow{1}{*}{\textbf{Elyadata}} & 67.50\textsubscript{±3.7} & 46.48\textsubscript{±10.1} & \textbf{57.09\textsubscript{±5.1}} & \textbf{50.57\textsubscript{±7.1}}\\
            \textcolor{blue}{BL I} & \multirow{1}{*}{Top 90\%} & 73.40\textsubscript{±6.1} & 60.67\textsubscript{±14.5} & 39.22\textsubscript{±14.6} & 45.09\textsubscript{±11.3}\\
            2& \multirow{1}{*}{\textbf{NLP\_DI}} & 71.88\textsubscript{±5.6} & 53.64\textsubscript{±10.2} & 37.42\textsubscript{±11.0} & 43.27\textsubscript{±9.4}\\
            \textcolor{blue}{BL II}& \multirow{1}{*}{Random} & 50.14\textsubscript{±1.6} & 30.43\textsubscript{±8.8} & 50.15\textsubscript{±2.1} & 37.15\textsubscript{±7.2}\\
            \textcolor{blue}{BL III} & \multirow{1}{*}{Top 1} & \textbf{73.42\textsubscript{±7.6}} & \textbf{76.82\textsubscript{±10.6}} & 17.77\textsubscript{±10.8} & 27.30\textsubscript{±12.6}\\

            3 & \multirow{1}{*}{\textbf{dzNlp}} & 71.38\textsubscript{±7.2} & 63.22\textsubscript{±10.7} & 12.87\textsubscript{±3.8} & 20.98\textsubscript{±5.2}\\

        \bottomrule
    \end{tabular}%
    }
    
    \caption{
    Systems' performance on the test set of Subtask~1. See Appendix §\ref{sec:detailed_subtask1_results} for a more detailed analysis.}
    \label{tab:subtask1_f1}
\end{table}
\begin{table}[tb]
\centering
\resizebox{0.65\columnwidth}{!}{%
    \begin{tabular}{clc}
    \toprule
    \textbf{Rank} & \textbf{System} &   \textbf{RMSE ($\downarrow$)} \\
    \midrule
    1 & \textbf{ASOS} & \textbf{0.1403} \\
    2 & \textbf{AlexUNLP-STM} & 0.1406 \\
    3 & \textbf{CUFE} & 0.2001 \\
    \cdashline{1-3}
    \textcolor{blue}{BL I} & Sentence ALDi & 0.2178 \\
    \textcolor{blue}{BL II} & Constant (0.67) & 0.2361 \\
    \textcolor{blue}{BL III} & Random & 0.3521 \\
    \bottomrule
    \end{tabular}%
    }
    \caption{Systems performance on Subtask~2 test set.}
    \label{tab:subtask2_RMSE_test}
\end{table}

\begin{table}[h]
\centering
\setlength{\tabcolsep}{4pt}
\resizebox{\columnwidth}{!}{%
\begin{tabular}{clccccc@{}}
\toprule
\multirow{2}{*}{\textbf{Rank}}& \multirow{2}{*}{\textbf{System}} & \multicolumn{4}{c}{\textbf{BLEU($\uparrow$)}}\\\cmidrule{3-7}
 & & \multicolumn{1}{c}{\textbf{Overall}} & \multicolumn{1}{c}{\textbf{Egy.}} & \multicolumn{1}{c}{\textbf{Emi.}} & \multicolumn{1}{c}{\textbf{Jor.}} & \multicolumn{1}{c}{\textbf{Pal.}} \\ \midrule
1 & \textbf{Arabic Train}             & \textbf{20.44}  & 16.57   & \textbf{23.38}  & \textbf{21.37}    & \textbf{20.62}     \\
2  & \textbf{Alson}                    & 17.46  & \textbf{16.76}   & 12.53  & 20.94    & 18.43      \\
3  &  \textbf{ASOS}                     & 17.13  & 14.82   & 19.39  & 15.80    & 18.38      \\
4 &  \textbf{CUFE}                     & 16.09  & 14.86   & 17.35  & 15.98    & 16.20      \\
5  &  \textbf{MBZUAI BLEU}              & 10.54  &  8.53   &  7.61  &  15.72    & 11.08      \\

6 &  \textbf{VBNN}                     &  9.24  & 8.62   &  6.30  & 11.79    &  10.54      \\ \cdashline{1-7}
\multicolumn{1}{l}{\textcolor{blue}{BL I}}            &   AraT5\textsubscript{v2}  &  6.87  &  9.38  &  4.61  & 4.90 &  8.13    \\ \cdashline{1-7}

7 &  \textbf{MBZUAI BADG}              &  2.78  &  3.03   &  1.74  &  3.91    &  2.48      \\\cdashline{1-7}
\multicolumn{1}{l}{\textcolor{blue}{BL II}}             & mT5 & 2.81  &  3.08  &  2.23  &  3.11  &  2.95    \\\cdashline{1-7}

\multicolumn{1}{l}{\textcolor{blue}{BL III}}        & AraBART &0.87  &  0.77  &  0.81  &  1.11  &  0.88     \\ \cdashline{1-7}
8 &  \textbf{Shaheen}                  & 0.00   & 0.00    &  0.00  &  0.00    &  0.00      \\ \bottomrule
\end{tabular}%
}
\caption{Performance of the systems on the test set of Subtask~3. Results are sorted by overall BLEU score.}
\label{tab:subtask3_bleu_test}
\end{table}

\subsection{Shared Task Results}\label{subsec:results}

\paragraph{Subtask~1} \textit{\textbf{Elyadata}} came first with a macro-averaged \texttt{F\textsubscript{1}} score of 50.57\%, being the only team to beat the \textit{{Top 90\%} baseline} model as per Table~\ref{tab:subtask1_f1}.

\paragraph{Subtask~2} \textit{\textbf{ASOS}}, the top-performing team, achieved the lowest RMSE of 0.1403, while \textit{\textbf{AlexUNLP-STM}} achieved a similar RMSE of 0.1406 coming second in the ranking. As shown in Table~\ref{tab:subtask2_RMSE_test}, all the teams managed to improve over our baselines, including systems trained on the AOC-ALDi dataset which has a different nature (comments on news not tweets) and was annotated based on less nuanced guidelines than ours.

\paragraph{Subtask~3} Table~\ref{tab:subtask3_bleu_test} shows the leaderboard of Subtask~3. 
\textit{\textbf{ArabicTrain}} 
won first place, achieving 
a BLEU score of $20.44$. We observe that \textit{six} teams outperform our best baseline 
on this subtask.

\begin{table*}[]
\setlength{\tabcolsep}{4pt}
\centering
\resizebox{0.95\textwidth}{!}{%
\begin{tabular}{@{}llcccccccccccc@{}}
\toprule
\multirow{2}{*}{\rotatebox[origin=c]{90}{\textbf{Task}}} & \multicolumn{1}{c}{\multirow{2}{*}{\textbf{Team}}} & \multirow{2}{*}{\textbf{Metric}} & \multicolumn{2}{c}{\textbf{Features}}                            & \multicolumn{9}{c}{\textbf{Techniques}}     \\ \cmidrule(l){4-5} \cmidrule(l){6-14} 
 & & & \textbf{N-gram} & \textbf{TFIDF} & \textbf{C-ML} & \textbf{NNs} & \textbf{PLM} & \textbf{LLM} & \textbf{Ensemble} & \textbf{Adapter} & \textbf{Post-Poc.} & \textbf{Cont. L.} & \textbf{D-Aug.}\\ \midrule
\multirow{3}{*}{\rotatebox[origin=c]{90}{1}} & Elyadata                                                               & 50.57                        &                            &                            &                                  & \checkmark                              & \checkmark                      &                         & \checkmark                           &                             &                                &                                      &                               \\
& NLP\_DI                                                                 & 43.27                        &                            &                            &                                  & \checkmark                              & \checkmark                      &                         & \checkmark                           &                             & \checkmark                             &                                      &                               \\
& dzNLP                                                                  & 20.98                        & \checkmark                         & \checkmark                         & \checkmark                               &                                 &                         &                         & \checkmark                           &                             &                                &                                      &                               \\ \midrule
\multirow{2}{*}{\rotatebox[origin=c]{90}{2}} & ASOS                                                                    & 0.1403                       &                            &                            &                                  & \checkmark                              & \checkmark                      &                         & \checkmark                           &                             &                                &                                      & \checkmark                            \\
& AlexUNLP-STM                                                           & 0.1406                       &                            &                            &                                  & \checkmark                              & \checkmark                      &                         & \checkmark                           &                             &                                & \checkmark                                   &                               \\ \midrule
\multirow{4}{*}{\rotatebox[origin=c]{90}{3}} & Arabic Train                                                            & 20.44                        &                            &                            &                                  &                                 &                         & \checkmark                      &                              & \checkmark                          &                                &                                      & \checkmark                            \\
& Alson                                                                  & 17.46                        &                            &                            &                                  &                                 & \checkmark                      & \checkmark                      &                              &                             &                                &                                      & \checkmark                            \\
& ASOS                                                                   & 17.13                        &                            &                            &                                  &                                 & \checkmark                      & \checkmark                      &                              &                             &                                &                                      & \checkmark                           \\ 
& CUFE                                                                   & 16.09                        &                            &                            &                                  &                                 & \checkmark                      & \checkmark                      &                              & \checkmark                          &                                &                                      &                               \\\bottomrule
\end{tabular}
}
\caption{Summary of approaches used by participating teams NADI 2024 shared task. Teams are sorted by their performance on the official metric of each subtask. \textit{C-ML} (Classifcal ML) indicates any non-neural machine learning methods such as naive Bayes and support vector machines. The term \textit{NNs} refers to any model based on neural networks (e.g. RNN, CNN, and Transformer) trained from scratch. \textit{PLM} refers to neural networks pretrained with unlabeled data such as MARBERT and has less than 1B parameters. \textit{LLM} refers to neural networks containing more than 1B parameters. 
Approaches also included contrastive loss (\textit{Cont. L.}) and data augmentation (\textit{D-Aug.})
}
\label{table:approaches_summary}
\end{table*}

\subsection{General Description of Submitted Systems}\label{subsec:papers-desc}
A summary of approaches employed by the various teams is provided in Table~\ref{table:approaches_summary}. We briefly describe the top systems for each subtask here.
\paragraph{Subtask~1} The winning team, \textit{\textbf{Elyadata}}, extracted dialectal vocabularies from the training data, and used them to augment the labels of the single-label training dataset. They then used a max pooling layer to merge the predictions of a MARBERT-based ensemble model forming an array of logit predictions. Lastly, they optimized a threshold using the development set to convert the logits into multi-label predictions.

\paragraph{Subtask~2}\textit{\textbf{ASOS}} fine-tuned a regression head of multiple layers on top of MARBERT's [CLS] embedding. \textit{\textbf{AlexUNLP-STM}} used the median of an ensemble of regression heads with sigmoid activation on top of AraBERT, trained to minimize contrastive and RMSE losses. 
Noticeably, their model's performance dropped when non-Arabic letters were discarded. We observed that code-switching affected the annotators' ALDi judgments differently, which is in-line with the team's 
justification for the performance drop.

%
\paragraph{Subtask~3}The winning team, \textit{\textbf{Arabic Train}}, utilized samples from MADAR~\cite{bouamor-etal-2019-madar} training set as the one-shot example to prompt LLM Jais~\cite{sengupta2023jais} for translating Arabic dialects to MSA. Team \textit{\textbf{Alson}} exploited ChatGPT to generate parallel data for translating Jordanian and Palestinian dialects to MSA and then fine-tuned AraT5 with generated samples and MADAR dataset. 
 
\section{Discussion}

\begin{figure}[tb]
    \centering
    \includegraphics[width=\columnwidth]{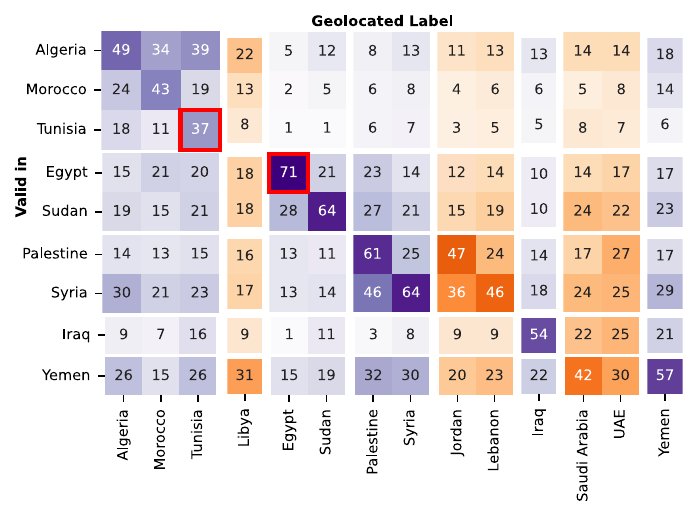}
    \caption{The number of DA samples valid in the annotators' country-level dialects (rows) across the 14 countries to which the samples are geolocated (columns). Each row represents the distribution of the geolocated labels for the sentences valid in the row's country-level dialect. \colorbox{orange}{Orange columns} 
    indicate the 
    countries not represented by the annotators. 
    The max cell value is 75.}
    \label{fig:geolocation_accuracy}
\end{figure}

\paragraph{Precision of Geolocated Labels} 
Although geolocation can alleviate the need for manually annotating the samples~\cite{abdul-mageed-etal-2021-nadi}, it can be error-prone~\cite{abdul-mageed-etal-2020-nadi,abdelali-etal-2021-qadi}.  
For the 1,050 DA samples of the development and test set, we can estimate the precision of the geolocated labels by comparing them to the manual validity labels as demonstrated in Figure~\ref{fig:geolocation_accuracy}. Based on this method, we find that the precision of the geolocated labels could be as high as 94.6\% (71 out of 75 samples) for Egypt, and as low as 49.3\% (37/75) for Tunisia. 


\paragraph{Impact of Named Entities} 
ADI models, trained on single-label data, can make spurious connections between named entities (e.g.: specific locations) and country-level labels \cite{abdul-mageed-etal-2020-toward}. In \texttt{NADI-2021-TWT} for example, 52 samples out of the 66 mentioning \begin{footnotesize}\AR{لبنان}\end{footnotesize} (Lebanon) are geolocated to and labeled as \textit{Lebanon}. Such spurious connections might be the reason why the following n-grams \begin{footnotesize}\AR{عراق}\end{footnotesize}, \begin{footnotesize}\AR{لبنان}\end{footnotesize}, \begin{footnotesize}\AR{تونس}\end{footnotesize}, \begin{footnotesize}\AR{يمن}\end{footnotesize} are among the most discriminative for the dialects of \textit{Iraq}, \textit{Lebanon}, \textit{Tunisia}, and \textit{Yemen}, respectively \cite{aalabdulsalam-2022-squ}. Manual annotation can alleviate this limitation. 


\subsection{Lessons Learned}
We share our reflections on the creation of evaluation datasets for Subtasks 1 and 2 as per~§\ref{sec:datasets}.


\paragraph{Subtask~1 Complexity} Previous research asking Arabic speakers to check the validity of sentences in their native dialects (See Table~\ref{tab:validity_wordings}) reported moderate to high agreement between the annotators (only two per country) for \textit{most} of the considered regional-level and country-level dialects. Unlike previous works, we recruited three annotators per country and asked them to judge all samples, rather than those geolocated to their own respective countries. Therefore, our annotation task is possibly harder than previous ones, which is reflected in the IAA scores in Table~\ref{tab:IAA_agg}.

\paragraph{Subtask~1 Labels} From a task design perspective, we observe that the frequency of usage of the \textit{Maybe (Not sure)} label varies across annotators. For this reason, including this particular label (rather than 
using a binary \textit{Valid/Not Valid} setup), needs to be further investigated to understand its implications on the aggregated validity labels.

\paragraph{Annotation Quality Monitoring} Two authors who are speakers of Egyptian Arabic were responsible for monitoring the quality of the annotations, providing feedback, and marking the samples with high disagreement for reannotation. We believe that having dialect leads who are native speakers of the different Arabic dialects would allow for better monitoring of the annotation process. We hope that our shared task will inspire future collaborative research to extend the labels of our evaluation dataset to include more country-level dialects.


\section{Conclusion}\label{sec:conclusion}
This year, we organized NADI 2024, the fifth edition of the shared task, having three subtasks: multi-label dialect identification (MDID), Arabic level of dialectness (ALDi) estimation, and DA-to-MSA machine translation. We had 51 registered teams, out of which 12 submitted their systems' predictions with eight accepted system description papers. The results indicate that there is still room for improvement across the various tasks. In the future, we intend to cover more Arabic dialects in NADI and propose novel ways of modeling that involve the use of large language models. 

\section*{Limitations}\label{sec:limis}

Our work has a number of limitations, as follows:

\begin{itemize}
\item This edition of NADI focused on only $10$ country-level dialects for Subtasks 1 and 2. This is due to challenges with recruiting annotators as well as the lack of high-quality datasets for countries such as \textit{Comoros},  \textit{Djibouti}, \textit{Mauritania}, and \textit{Somalia}.

\item NADI continues to use short texts for the Arabic dialects. That is, due to the shortage of dialectal data from other sources, we depend on posts from Twitter. Although these data have thus far empowered the development of effective dialect identification models, it is desirable to afford data from other domains that have longer texts. This will allow the development of more widely applicable models.

\item The label aggregation techniques (See §\ref{sec:label_agg}) used for the evaluation sets of Subtask~1,~2 attempts to reduce the impact of the few inevitable inaccurate annotations. However, they could also inhibit interannotator disagreement that is caused by having different perceptions (i.e., what sentences are valid in their country-level dialects, or what the level of dialectness of sentences are) \cite{ovesdotter-alm-2011-subjective, davani-etal-2022-dealing}.

\item Our machine translation subtask focuses only on four dialects without offering a training dataset. Modern MT systems need much larger data to perform well. Again, in spite of our best efforts, parallel datasets involving dialects remain limited.

\item Due to limited resources, we were able to provide only a single reference annotation for Subtask~3 test samples. We acknowledge that machine translation 
requires multiple evaluation references to ensure a more reliable assessment.  

\item We acknowledge that the BLEU score for evaluating machine translation output has its limitations~\cite{popovic-2017-chrf,kocmi-etal-2021-ship,rei-etal-2022-comet}. We expect that using more diverse metrics, such as ChrF~\cite{popovic-2017-chrf} and COMET~\cite{rei-etal-2022-comet}, can enhance the reliability of evaluation results.
\end{itemize}

\section*{Ethical Considerations}\label{sec:ethics}
The NADI-2024 Subtask~1,~2 datasets are sourced from the public domain (i.e., X former Twitter), with user personal information and identity carefully concealed. Similarly, the NADI-2024 Subtask~3 dataset is manually created. Again, we take meticulous measures to remove user identities and personal information from our datasets. As a result, we have minimal concerns about the retrieval of personal information from our data. However, it is crucial to acknowledge that the datasets we collect to construct NADI-2024 Subtask~1,~2 may contain potentially harmful content. Additionally, during model evaluation, there is a possibility of exposure to biases that could unintentionally generate problematic content.

Finally, we note that the annotation process we followed for creating the evaluation dataset of the first two subtasks was approved by the research ethics committee of the University of Edinburgh, School of Informatics with reference number 839548.

\section*{Acknowledgments}
Muhammad Abdul-Mageed acknowledges support from Canada Research Chairs (CRC), the Natural Sciences and Engineering Research Council of Canada (NSERC; RGPIN-2018-04267), the Social Sciences and Humanities Research Council of Canada (SSHRC; 435-2018-0576; 895-2020-1004; 895-2021-1008), Canadian Foundation for Innovation (CFI; 37771), Digital Research Alliance of Canada,\footnote{\href{https://alliancecan.ca}{https://alliancecan.ca}} and UBC ARC-Sockeye.

Amr Keleg's work is supported by the UKRI Centre for Doctoral Training in Natural Language Processing, funded by the UKRI (grant EP/S022481/1) and the University of Edinburgh, School of Informatics.

\bibliography{anthology,custom,dlnlp,NADI2023-paper,team_referances}

\appendix
\clearpage
\appendixpage
\addappheadtotoc
\setcounter{table}{0}
\setcounter{figure}{0}
\renewcommand{\thetable}{\Alph{section}\arabic{table}}
\renewcommand{\thefigure}{\Alph{section}\arabic{figure}}

\section{Design of the Annotation Guidelines}\label{sec:annotation_guidelines}

In this year's edition of NADI, we introduced two new subtasks, MDID, and ALDi Estimation. As explained in §\ref{sec:datasets}, we annotated 1,120 tweets for both subtasks to build the development and test splits.

\paragraph{Subtask~1}
There has been multiple attempts asking annotators to check if sentences are written in their Arabic dialect. This was done mainly to validate dialect labels that are automatically assigned using geolocating methods/distinctive dialectal cues \cite{alsarsour-etal-2018-dart,abdelali-etal-2021-qadi,Althobaiti_2022} or to perform error analysis for the predictions of DI systems \cite{keleg-magdy-2023-arabic}. 
Arabic speakers have different perceptions of their country-level dialects that depend on their backgrounds and exposure to different speaking communities. Such differences could impact their understanding of the validity of sentences in their country-level dialects. Moreover, previous wordings used to check the validity of these sentences shown in Table~\ref{tab:validity_wordings} were used to validate the labels for sentences that are more probable to be in the annotator's native dialect (i.e., not in another dialect nor in MSA).

\begin{figure}[tb]
    \centering
        \centering
        \fbox{\includegraphics[trim={0 4cm 0 2cm}, clip, width=\columnwidth]{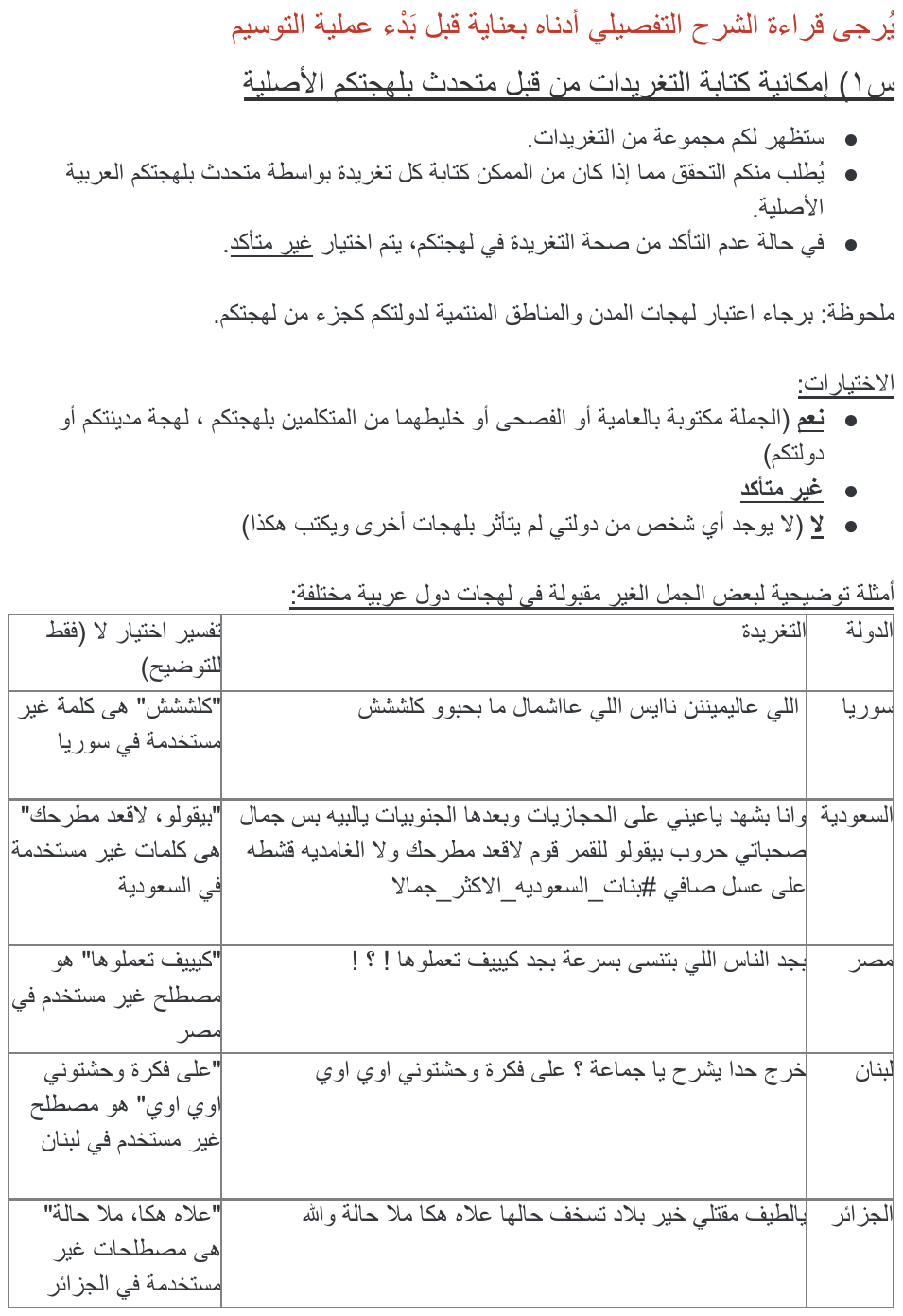}}
    \caption{The guidelines used for annotating the validity of the sentences (Subtask~1 - MDID).}
    \label{fig:validity_guidelines}
\end{figure}

Conversely, we asked the annotators to label 1120, uniformly representing 14 country-level dialects. Moreover, we are interested in checking the validity of the sentences in any of the dialects spoken in the annotator's country of origin, and not just in their native Arabic dialect. Consequently, we used the wording mentioned in §\ref{sec:annotation_process}, in addition to providing some examples as shown in Figure~\ref{fig:validity_guidelines}.
\begin{table}[!ht]
    \scriptsize
    \centering
    \begin{tabular}{p{0.8\columnwidth}}
         \multicolumn{1}{c}{\textbf{Wording}}  \\
         \midrule
         \cite{alsarsour-etal-2018-dart}\\
         Asked annotators to label each tweet as either in:\\(their native dialect, MSA, or other).\\
         \midrule
         \cite{abdelali-etal-2021-qadi}\\
         \textit{Is this tweet consistent with the dialect spoken in your country?} (Yes, No).\\
         \midrule
         \cite{Althobaiti_2022}\\
         \textit{Is this sentence written in \texttt{Dialect} dialect?} (Yes, No);\\ \texttt{Dialect} is the demonymic form of the annotator's country. \\
         \midrule
         \cite{keleg-magdy-2023-arabic}\\
         \textit{Is this sentence valid in your dialect?} (Yes, Not Sure, No).\\
        \bottomrule
    \end{tabular}
    \caption{Previous wordings for checking the validity of sentences in an Arabic dialect.}
    \label{tab:validity_wordings}
\end{table}

\begin{figure*}[ph]
    \centering
    \includegraphics[width=0.9\textwidth]{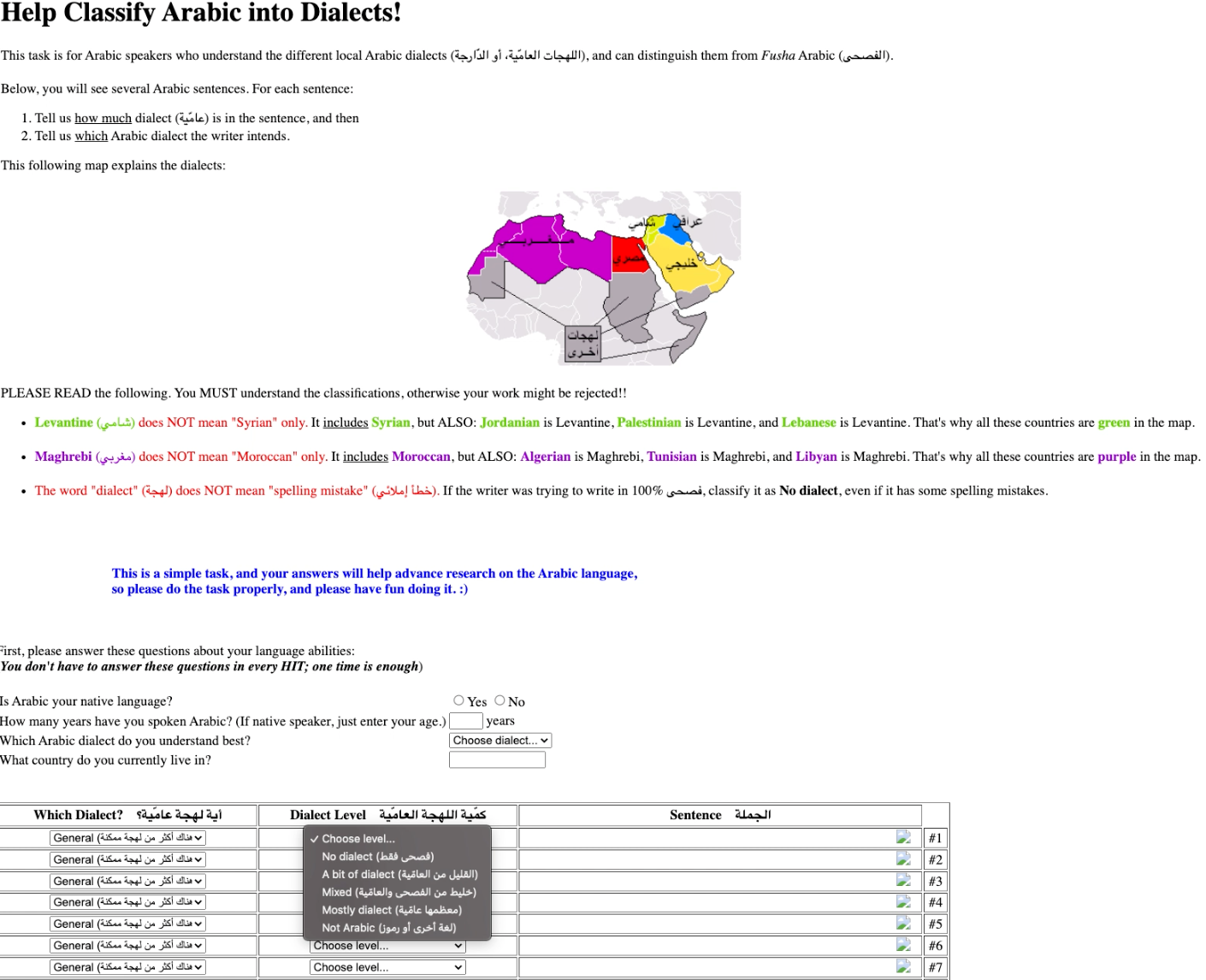}
    \caption{A screenshot of the annotation interface of the AOC dataset \cite{zaidan-callison-burch-2011-arabic}.}
    \label{fig:AOC_interface}

    \vspace{0.5cm}

   \begin{subfigure}[b]{0.45\textwidth}
        \centering
        \fbox{\includegraphics[trim={0 9cm 0 2cm}, clip, width=\textwidth]{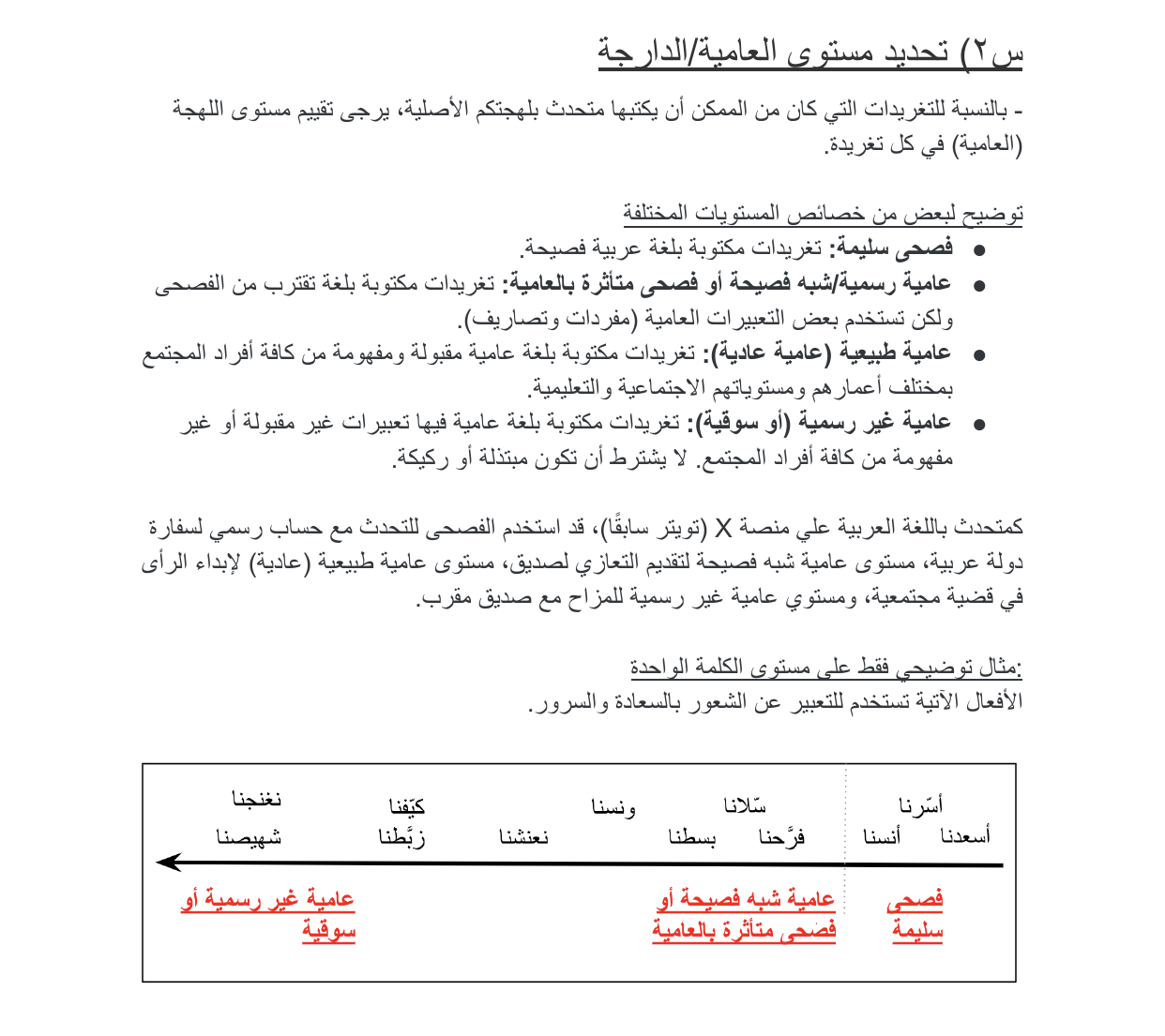}}
        \caption{The guidelines for the ALDi Estimation subtask.}
   \end{subfigure}
   \hfill
   \begin{subfigure}[b]{0.45\textwidth}
        \centering
        \fbox{\includegraphics[trim={0 9cm 0 3cm}, clip, width=\textwidth]{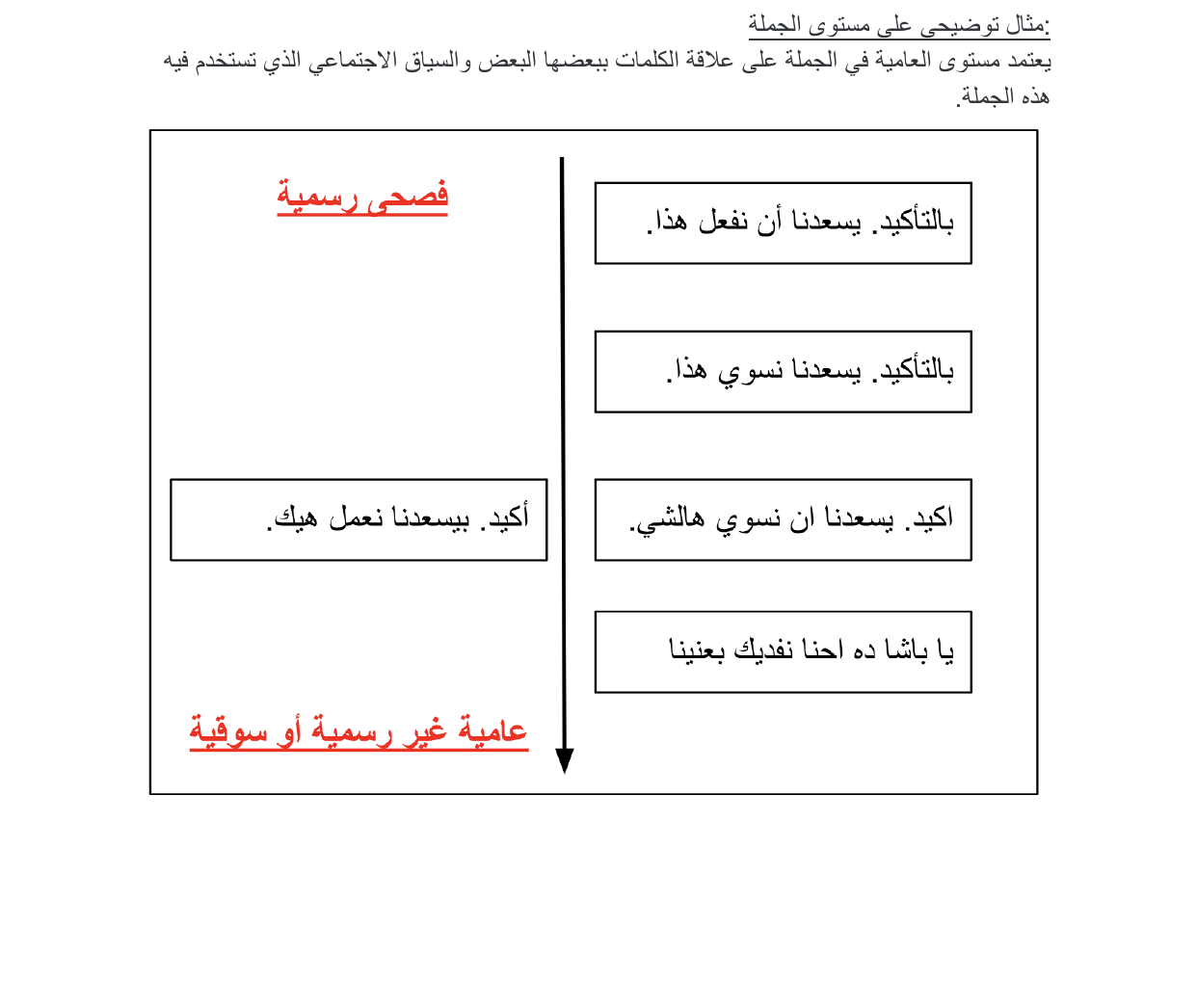}}
        \caption{An example of different-ALDi variants of a sentence.}
    \end{subfigure}

    \caption{Screenshots of the guidelines that were provided to the annotators to determine the ordinal ALDi level (Subtask~2) for the 1120 sentences of the development and test sets.}
    \label{fig:ALDi_annotation_guidelines}
\end{figure*}

\paragraph{Subtask~2} We follow \citet{zaidan-callison-burch-2011-arabic}'s setup in which they asked the annotators to assign a discrete ALDi level to each sentence as shown in Figure~\ref{fig:AOC_interface}. In their setup, we noticed that Level 3 (\textit{Mostly Dialectal} (\begin{scriptsize}\AR{معظمها عامية}\end{scriptsize})) could not fully separate between sentences having a word perceived as highly dialectal, and sentences having a majority of dialectal words that are not perceived as highly dialectal on the word level. Therefore, we provide descriptive labels to the 4 levels, and short descriptions for the sentences of each label (See~\ref{sec:annotation_process} for the English translation of these labels/descriptions). Moreover, we use two examples to further explain the concept of ALDi on the word and sentence level, as per Figure~\ref{fig:ALDi_annotation_guidelines}. This would allow for better separation between the different ALDi levels, and higher IAA scores. 

\section{Evaluation Data Annotation Process}\label{sec:annotation_process_details}
As summarized in §\ref{sec:annotation_process}, we asked the annotators to complete a short onboarding task before joining the 5 main annotation tasks. In this section, we provide further details about the annotation process.

\paragraph{Onboarding Tasks}
QADI's test set \cite{abdelali-etal-2021-qadi} has 3,303 tweets geolocated to 18 different Arab countries (including the 14 countries represented in the samples of our dataset). The geolocated label for each tweet was then validated by a native speaker from that country, who checked if the ``tweet is consistent with the dialect spoken in their country".
Additionally, the test set has 200 tweets automatically classified as written in MSA.

In order to get the annotators acquainted with our annotation guidelines shown in Figure~\ref{fig:validity_guidelines}, and Figure~\ref{fig:ALDi_annotation_guidelines}, we asked them to label 35 tweets from QADI as an initial onboarding task. Each country's onboarding task had 10 tweets labeled as consistent with the dialect(s) spoken in the country according to QADI's annotations and 5 MSA samples.\footnote{We noticed that some non-MSA samples of QADI were flagged as being in MSA by our in-house MSA/DA classifier, so we relied on the predictions of our model for considering a sample as MSA.} We included MSA samples to ensure that the onboarding tasks contain tweets of potentially different levels of dialectness. Additionally, we had 2 DA samples from 10 other country-level dialects, that would act as negatives for the first subtask (i.e., some of these samples are expected not to be valid in the considered country of the onboarding task).
For each country's onboarding task, the composing samples were randomly shuffled. The annotators were not given information about the samples' geolocated labels or their distribution across the different countries/labels.

\paragraph{Quality Assurance and Feedback} For each country's onboarding task, and thanks to the labels from QADI, we could perform two automatic checks for assessing the quality of the annotations:
\begin{itemize}
    \item \textbf{Check (1)} The 10 samples geolocated to an annotator's country are expected to (a) be labeled as valid, and (b) with an ALDi level~>~0.
    \item \textbf{Check (2)} The 5 MSA samples are expected to be marked as valid by all the annotators, with Level (0) \textit{Sound MSA} as their ALDi.
\end{itemize}

\begin{sidewaystable*}[tbhp]
\scriptsize
\aboverulesep=0ex 
\belowrulesep=0ex 
\rule{0pt}{1.1EM}
\centering
\begin{tabular}{l|ccccccccccccccc}
\multirow{3}{*}{\textbf{Annotator \#}}& \multicolumn{13}{c}{\textbf{Sample Geolocated Label}}\\
& \textbf{Algeria} & \textbf{Egypt} & \textbf{Iraq} & \textbf{Jordan} & \textbf{Lebanon} & \textbf{Libya} & \textbf{Morocco} & \textbf{Palestine} & \textbf{Saudi Arabia} & \textbf{Sudan} & \textbf{Syria} & \textbf{Tunisia} & \textbf{UAE} & \textbf{Yemen} & \textbf{MSA} \\
& \textbf{(y/m/n)} & \textbf{(y/m/n)} & \textbf{(y/m/n)} & \textbf{(y/m/n)} & \textbf{(y/m/n)} & \textbf{(y/m/n)} & \textbf{(y/m/n)} & \textbf{(y/m/n)} & \textbf{(y/m/n)} & \textbf{(y/m/n)} & \textbf{(y/m/n)} & \textbf{(y/m/n)} & \textbf{(y/m/n)} & \textbf{(y/m/n)} & \textbf{(y/m/n)} \\
\midrule
Algeria (A) & \textbf{6}/0/4 * &  1/0/\textbf{1} &  0/0/\textbf{2} & 0/0/\textbf{2} & 0/0/\textbf{2} &     - &  0/0/\textbf{2} & 0/0/\textbf{2} & 0/0/\textbf{2} & 0/0/\textbf{2} & 1/0/\textbf{1} & 1/1/\textbf{0} &   - &               - & \textbf{5}/0/0 \\
Algeria (B) & \textbf{7}/0/3 * &  1/0/\textbf{1} &  0/0/\textbf{2} & 0/0/\textbf{2} & 0/0/\textbf{2} &     - &  1/0/\textbf{1} & 0/0/\textbf{2} & 0/0/\textbf{2} & 1/0/\textbf{1} & 1/0/\textbf{1} & 2/0/\textbf{0} &   - &               - & \textbf{5}/0/0 \\
Algeria (C) & \textbf{5}/0/5 * &  0/0/\textbf{2} &  0/0/\textbf{2} & 0/0/\textbf{2} & 0/0/\textbf{2} &     - &  0/0/\textbf{2} & 0/0/\textbf{2} & 0/0/\textbf{2} & 0/0/\textbf{2} & 1/0/\textbf{1} & 1/0/\textbf{1} &   - &               - & \textbf{0}/0/5 * \\
\midrule
Egypt (A) & 0/0/\textbf{2} &  \textbf{8}/0/2 &  0/0/\textbf{2} & 0/0/\textbf{2} & 0/0/\textbf{2} &     - &  0/0/\textbf{2} & 0/0/\textbf{2} & 0/0/\textbf{2} & 1/0/\textbf{1} & 0/0/\textbf{2} & 0/0/\textbf{2} &   - &               - & \textbf{5}/0/0 \\
Egypt (B) & 0/1/\textbf{1} & \textbf{10}/0/0 &  1/0/\textbf{1} & 0/0/\textbf{2} & 0/0/\textbf{2} &     - &  0/0/\textbf{2} & 0/0/\textbf{2} & 0/0/\textbf{2} & 0/0/\textbf{2} & 0/0/\textbf{2} & 0/0/\textbf{2} &   - &               - & \textbf{5}/0/0 \\
Egypt (C) & 0/0/\textbf{2} & \textbf{10}/0/0 &  0/0/\textbf{2} & 0/0/\textbf{2} & 0/0/\textbf{2} &     - &  0/0/\textbf{2} & 0/1/\textbf{1} & 0/0/\textbf{2} & 1/0/\textbf{1} & 0/0/\textbf{2} & 0/0/\textbf{2} &   - &               - & \textbf{4}/0/1 \\
\midrule
Iraq (A) & 1/1/\textbf{0} &  1/0/\textbf{1} &  \textbf{6}/1/3 & 0/0/\textbf{2} & 0/0/\textbf{2} &     - &  0/0/\textbf{2} & 0/0/\textbf{2} & 0/0/\textbf{2} & 0/0/\textbf{2} & 0/0/\textbf{2} & 0/0/\textbf{2} &   - &               - & \textbf{5}/0/0 \\
Iraq (B) & 0/2/\textbf{0} &  0/1/\textbf{1} &  \textbf{7}/3/0 & 0/0/\textbf{2} & 0/0/\textbf{2} &     - &  0/0/\textbf{2} & 0/0/\textbf{2} & 0/0/\textbf{2} & 0/0/\textbf{2} & 0/0/\textbf{2} & 0/0/\textbf{2} &   - &               - & \textbf{3}/2/0 \\
Iraq (C) & 2/0/\textbf{0} &  1/0/\textbf{1} & \textbf{10}/0/0 & 0/0/\textbf{2} & 0/0/\textbf{2} &     - &  0/0/\textbf{2} & 0/0/\textbf{2} & 0/0/\textbf{2} & 0/0/\textbf{2} & 0/0/\textbf{2} & 0/1/\textbf{1} &   - &               - & \textbf{5}/0/0 \\
\midrule
Morocco (A) & 2/0/\textbf{0} &  2/0/\textbf{0} &  1/0/\textbf{1} & 2/0/\textbf{0} & 2/0/\textbf{0} &     - & \textbf{10}/0/0 & 1/1/\textbf{0} & 1/1/\textbf{0} & 2/0/\textbf{0} & 2/0/\textbf{0} & 2/0/\textbf{0} &   - &               - & \textbf{5}/0/0 \\
Morocco (B) & 2/0/\textbf{0} &  2/0/\textbf{0} &  2/0/\textbf{0} & 2/0/\textbf{0} & 2/0/\textbf{0} &     - & \textbf{10}/0/0 & 2/0/\textbf{0} & 2/0/\textbf{0} & 1/1/\textbf{0} & 2/0/\textbf{0} & 2/0/\textbf{0} &   - &               - & \textbf{5}/0/0 \\
Morocco (C) & 1/0/\textbf{1} &  0/0/\textbf{2} &  0/0/\textbf{2} & 0/0/\textbf{2} & 0/0/\textbf{2} &     - &  \textbf{8}/0/2 & 0/0/\textbf{2} & 0/0/\textbf{2} & 0/0/\textbf{2} & 1/0/\textbf{1} & 1/0/\textbf{1} &   - &               - & \textbf{4}/0/1 \\
\midrule
Palestine (A) & 1/0/\textbf{1} &  1/0/\textbf{1} &  0/0/\textbf{2} & 1/0/\textbf{1} & 1/0/\textbf{1} &     - &  0/0/\textbf{2} & \textbf{8}/0/2 & 0/0/\textbf{2} & 0/0/\textbf{2} & 1/0/\textbf{1} & 1/0/\textbf{1} &   - &               - & \textbf{4}/0/1 \\
Palestine (B) & 2/0/\textbf{0} &  2/0/\textbf{0} &  1/0/\textbf{1} & 2/0/\textbf{0} & 2/0/\textbf{0} &     - &  1/0/\textbf{1} & \textbf{7}/0/3 & 0/0/\textbf{2} & 0/0/\textbf{2} & 2/0/\textbf{0} & 1/0/\textbf{1} &   - &               - & \textbf{4}/0/1 \\
Palestine (C) & 1/0/\textbf{1} &  1/1/\textbf{0} &  0/0/\textbf{2} & 1/0/\textbf{1} & 2/0/\textbf{0} &     - &  0/0/\textbf{2} & \textbf{9}/1/0 & 0/0/\textbf{2} & 1/0/\textbf{1} & 1/0/\textbf{1} & 1/0/\textbf{1} &   - &               - & \textbf{5}/0/0 \\
\midrule
Sudan (A) & 0/1/\textbf{1} &  1/0/\textbf{1} &  0/0/\textbf{2} & 0/0/\textbf{2} & 0/0/\textbf{2} &     - &  0/0/\textbf{2} & 0/0/\textbf{2} & 0/0/\textbf{2} & \textbf{9}/1/0 & 0/0/\textbf{2} & 0/0/\textbf{2} &   - &               - & \textbf{4}/0/1 \\
Sudan (B) & 2/0/\textbf{0} &  0/0/\textbf{2} &  0/0/\textbf{2} & 0/0/\textbf{2} & 0/0/\textbf{2} &     - &  0/0/\textbf{2} & 0/0/\textbf{2} & 0/0/\textbf{2} & \textbf{9}/0/1 & 0/0/\textbf{2} & 0/0/\textbf{2} &   - &               - & \textbf{5}/0/0 \\
Sudan (C) & 2/0/\textbf{0} &  0/0/\textbf{2} &  0/1/\textbf{1} & 0/0/\textbf{2} & 0/0/\textbf{2} &     - &  0/0/\textbf{2} & 0/0/\textbf{2} & 0/0/\textbf{2} & \textbf{9}/1/0 & 1/0/\textbf{1} & 1/0/\textbf{1} &   - &               - & \textbf{5}/0/0 \\
\midrule
Syria (A) & 0/0/\textbf{2} &  1/0/\textbf{1} &  0/0/\textbf{2} & 1/0/\textbf{1} & 2/0/\textbf{0} &     - &  0/0/\textbf{2} & 0/0/\textbf{2} & 0/0/\textbf{2} & 0/0/\textbf{2} & \textbf{9}/0/1 & 0/0/\textbf{2} &   - &               - & \textbf{5}/0/0 \\
Syria (B) & 2/0/\textbf{0} &  1/0/\textbf{1} &  1/0/\textbf{1} & 1/0/\textbf{1} & 1/1/\textbf{0} &     - &  0/1/\textbf{1} & 0/0/\textbf{2} & 0/0/\textbf{2} & 0/0/\textbf{2} & \textbf{9}/1/0 & 1/1/\textbf{0} &   - &               - & \textbf{5}/0/0 \\
Syria (C) & 0/1/\textbf{1} &  1/0/\textbf{1} &  0/0/\textbf{2} & 0/0/\textbf{2} & 1/0/\textbf{1} &     - &  0/0/\textbf{2} & 0/0/\textbf{2} & 0/0/\textbf{2} & 0/0/\textbf{2} & \textbf{9}/1/0 & 0/0/\textbf{2} &   - &               - & \textbf{1}/4/0 * \\
\midrule
Tunisia (A) & 1/0/\textbf{1} &  0/0/\textbf{2} &  0/0/\textbf{2} & 0/0/\textbf{2} & 0/0/\textbf{2} &     - &  0/0/\textbf{2} & 1/0/\textbf{1} & 0/0/\textbf{2} & 0/0/\textbf{2} & 1/0/\textbf{1} & \textbf{9}/0/1 &   - &               - & \textbf{2}/0/3 * \\
Tunisia (B) & 1/0/\textbf{1} &  1/0/\textbf{1} &  0/0/\textbf{2} & 0/0/\textbf{2} & 0/0/\textbf{2} &     - &  0/0/\textbf{2} & 0/0/\textbf{2} & 0/0/\textbf{2} & 0/0/\textbf{2} & 0/0/\textbf{2} & \textbf{9}/0/1 &   - &               - & \textbf{5}/0/0 \\
Tunisia (C) & 1/0/\textbf{1} &  0/0/\textbf{2} &  0/0/\textbf{2} & 0/0/\textbf{2} & 0/0/\textbf{2} &     - &  0/0/\textbf{2} & 0/0/\textbf{2} & 0/0/\textbf{2} & 0/0/\textbf{2} & 1/0/\textbf{1} & \textbf{8}/1/1 &   - &               - & \textbf{4}/1/0 \\
\midrule
Yemen (A) & 1/1/\textbf{0} &  0/0/\textbf{2} &  0/0/\textbf{2} & 0/0/\textbf{2} & 0/0/\textbf{2} &     - &  0/0/\textbf{2} & 0/1/\textbf{1} & 0/0/\textbf{2} & 0/0/\textbf{2} & 1/0/\textbf{1} & 1/0/\textbf{1} &   - &  \textbf{8}/1/1 & \textbf{5}/0/0 \\
Yemen (B) & 2/0/\textbf{0} &  1/0/\textbf{1} &  0/0/\textbf{2} & 0/0/\textbf{2} & 0/0/\textbf{2} &     - &  0/0/\textbf{2} & 0/0/\textbf{2} & 1/0/\textbf{1} & 0/0/\textbf{2} & 1/0/\textbf{1} & 1/0/\textbf{1} &   - & \textbf{10}/0/0 & \textbf{4}/0/1 \\
Yemen (C) & 2/0/\textbf{0} &  1/0/\textbf{1} &  0/0/\textbf{2} & 0/0/\textbf{2} & 1/0/\textbf{1} &     - &  0/0/\textbf{2} & 0/0/\textbf{2} & 0/1/\textbf{1} & 0/0/\textbf{2} & 1/0/\textbf{1} & 0/1/\textbf{1} &   - &  \textbf{8}/1/1 & \textbf{5}/0/0 \\
\bottomrule
\end{tabular}
    \caption{The distribution of the validity labels for the samples of the onboarding tasks presented as the number of each of the following decisions (Yes/Maybe/No), split into columns according to QADI's geolocated label of the samples. \textbf{Note \#1:} The bolded value in each column represents the expected decision. \textbf{Note \#2}: We initially discarded Libya, UAE, and Yemen from our dataset, and thus the onboarding datasets of the other countries do not have samples from these three countries. \textbf{Note \#3}: We marked the unexpected patterns with *.}
    \label{tab:onboarding_validity_distribution}
\end{sidewaystable*}

On inspecting the annotators' performance on the onboarding task, we found that these automated checks are generally satisfied by the annotators, as per Table \ref{tab:onboarding_validity_distribution}. The checks also helped us identify any misinterpretation of the guidelines, and provide feedback to our annotators before labeling the samples of the main task. For instance, one of the Algerian annotators \textit{Algeria (C)} interpreted question (1) as identifying if the tweet matches how an Algerian typically writes, and consequently marked all the MSA samples as not valid in Algerian Arabic. For these sentences, he provided an alternative translation that sounds more natural to him (e.g., for the sample \begin{small}\AR{واحلى اخت انتي . . نفتخر و نعتز بيك اكيد ربي يحفظك}\end{small}, he provided the following alternative \begin{small}\AR{انتي اخت مخيرة .. نزوخ و نستعرف بيك ربي يحفظك}\end{small}). Another Syrian annotator \textit{Syria (C)} chose \textit{Maybe/ Not Sure} for 4 out of the 5 MSA samples, as she understood the first question as if the sentence could have \underline{\textit{only}} been written by a speaker of Syrian Arabic. Her explanation for her choice is: \textit{For the MSA sentences, it is hard to accurately identify the speaker's dialect, so I chose unsure}.

\begin{table*}[tb]
\renewcommand{\arraystretch}{1.3}
\tiny
\centering
\begin{tabular}{r}
\toprule
\multicolumn{1}{c}{\textbf{Samples from Algeria's Onboarding Task deemed invalid by the three Algerian annotators}} \\
\midrule
\RL{* ههههههههههههههههههههههههههههههههههههه ياي وبتصحح كمان مصدقة نفسها يابنتي مش هيعبرك برضو سايكو بجد هموت}\\
\RL{* بسم الله الرحمن راح نبدأ نعد واجيب البت دي تعدي معايا}\\
\RL{* عباس لسه ما فيه وسم لمظاهرة اليوم نزلو وقول وززززززع وخليه يوصل ترند عالمي}\\
\bottomrule
\end{tabular}
\caption{Three samples categorically annotated as invalid by the three Algerian annotators, yet are geolocated to Algeria according to QADI's test set.}
\label{tab:onboarding_Algerian_samples_labeled_as_not_valid}
\end{table*}

\begin{table*}[tb]

\begin{subtable}[c]{0.5\textwidth}
\scriptsize
\centering
    \begin{tabular}{lccccc}
    \toprule
    \multirow{2}{*}{\textbf{Country}} & \multicolumn{5}{c}{\textbf{Fleiss' Kappa ($\kappa$)}}\\
     &   \textbf{Batch 1} &   \textbf{Batch 2} &   \textbf{Batch 3} &    \textbf{Batch 4} &   \textbf{Batch 5} \\
    \midrule
    \multirow{2}{*}{Algeria} &  0.49 (59) & 0.49 (80) & 0.54 (65) &  0.51 (67) & 0.48 (59) \\
    &  0.58 (59) &         - &         - &  0.61 (70) & 0.59 (59) \\
    \midrule
    \multirow{2}{*}{Morocco} &  0.62 (34) & 0.42 (28) & 0.27 (49) &  0.36 (41) & 0.48 (43) \\
    &          - & 0.81 (53) &  0.5 (50) &  0.53 (46) & 0.62 (47) \\
    \midrule
    \multirow{2}{*}{Tunisia} &  0.43 (28) & 0.67 (47) & 0.64 (33) &  0.53 (31) & 0.46 (32) \\
    &  0.56 (41) &         - & 0.71 (40) &   0.7 (31) & 0.71 (30) \\
    \midrule
    \multirow{2}{*}{Egypt} &  0.58 (62) & 0.63 (69) & 0.56 (68) &  0.64 (81) & 0.69 (74) \\
    &   0.7 (61) &         - &         - &  0.74 (81) & 0.79 (74) \\
    \midrule
    \multirow{2}{*}{Sudan} &  0.57 (68) & 0.53 (76) & 0.58 (84) &  0.56 (81) & 0.67 (77) \\
    &  0.72 (74) &         - &         - &  0.74 (81) & 0.79 (78) \\
    \midrule
    \multirow{2}{*}{Palestine} &  0.4 (114) & 0.41 (59) & 0.52 (72) &  0.51 (61) & 0.54 (71) \\
    & 0.58 (111) &         - &         - &  0.69 (67) & 0.74 (66) \\
    \midrule
    \multirow{2}{*}{Syria} &  0.39 (83) & 0.49 (92) & 0.49 (87) & 0.59 (109) & 0.53 (90) \\
    &  0.56 (89) &         - &         - &          - & 0.57 (98) \\
    \midrule
    \multirow{2}{*}{Iraq} &  0.59 (58) & 0.52 (44) & 0.59 (50) &  0.61 (51) & 0.59 (53) \\
    &          - &         - &         - &  0.69 (62) & 0.64 (57) \\
    \midrule
    \multirow{2}{*}{Yemen} & 0.46 (101) & 0.57 (99) & 0.52 (81) &  0.45 (76) & 0.45 (94) \\
    & 0.55 (104) &         - &         - &          - &  0.5 (94) \\
    \bottomrule
    \end{tabular}

\subcaption{Subtask~1 - Validity Labels.}
\label{tab:kappa_scores_and_n_valid}
\end{subtable}
~
\begin{subtable}[c]{0.5\textwidth}
\scriptsize
\centering
    \begin{tabular}{ccccc}
    \toprule
    \multicolumn{5}{c}{\textbf{Krippendorff's Alpha ($\alpha$)}}\\
    \textbf{Batch 1} &   \textbf{Batch 2} &   \textbf{Batch 3} &    \textbf{Batch 4} &   \textbf{Batch 5} \\
    \midrule
    0.745 (59) & 0.615 (80) & 0.708 (65) &  0.513 (67) & 0.715 (59) \\
    0.663 (59) &          - &          - &  0.536 (70) & 0.666 (59) \\
    \midrule
    0.823 (34) & 0.691 (28) & 0.767 (49) &  0.768 (41) & 0.687 (43) \\
    - &  0.76 (53) & 0.742 (50) &  0.811 (46) & 0.742 (47) \\
    \midrule
    0.798 (28) & 0.738 (47) &   0.8 (33) &   0.71 (31) & 0.664 (32) \\
    0.787 (41) &          - & 0.808 (40) &  0.722 (31) & 0.698 (30) \\
    \midrule
    0.845 (62) & 0.828 (69) & 0.791 (68) &  0.791 (81) & 0.862 (74) \\
    0.796 (61) &          - &          - &  0.788 (81) &  0.82 (74) \\
    \midrule
    0.765 (68) & 0.537 (76) & 0.657 (84) &  0.696 (81) & 0.624 (77) \\
    0.746 (74) &          - &          - &  0.727 (81) & 0.643 (78) \\
    \midrule
    0.731 (114) & 0.752 (59) & 0.673 (72) &  0.633 (61) & 0.559 (71) \\
    0.645 (111) &          - &          - &  0.739 (67) & 0.573 (66) \\
    \midrule
    0.845 (83) & 0.709 (92) & 0.866 (87) & 0.751 (109) & 0.774 (90) \\
    0.829 (89) &          - &          - &           - & 0.796 (98) \\
    \midrule
    0.677 (58) & 0.684 (44) & 0.724 (50) &  0.733 (51) & 0.795 (53) \\
    - &          - &          - &  0.776 (62) & 0.816 (57) \\
    \midrule
    0.561 (101) & 0.495 (99) & 0.457 (81) &  0.568 (76) & 0.397 (94) \\
    0.498 (104) &          - &          - &           - & 0.433 (94) \\
    \bottomrule
    \end{tabular}
\subcaption{Subtask~2 - ALDi Labels.}
\label{tab:krip_scores_and_n_valid}
\end{subtable}

\caption{The detailed IAA scores for each of the 5 main annotation tasks, computed independently for each country's 3 annotators. The second line for each country represents the IAA scores after providing feedback to the annotators and asking them to reannotate the samples of high disagreement. \textbf{Note:} The number of sentences valid in each country-level dialect after applying majority voting is shown between (brackets).}
\label{tab:IAA_detailed}
\end{table*}

Moreover, we noticed that for the onboarding task of Algeria, 3 samples geolocated to Algeria (Table \ref{tab:onboarding_Algerian_samples_labeled_as_not_valid}), are labeled as not valid in Algerian Arabic by all the three Algerian annotators. 
Given that (a) the members of our team responsible for the annotation process are native speakers of Egyptian Arabic only, and (b) the labels for the main task are not validated (i.e., assigned based on geolocation only), we could not judge whether these are errors in the labels provided by our annotators or if QADI's validated labels for these samples were incorrect. Therefore, we decided not to use these checks as automatic measures for accepting or rejecting the labels provided by the annotators for the main task batches, and resorted to manually inspecting the annotations by the end of each annotation batch, as elaborated next.

\paragraph{Main Task Batches}
Following the onboarding task, we invited the annotators to label the task's data, split into 5 batches, of 224 samples each. We ran the annotation batches over 5 weeks (1 batch per week), to ensure a higher annotation quality.

By the end of each batch, and as done for the onboarding task, we used the 2 aforementioned checks to inspect the quality of the annotations.
Moreover, we compared the labels provided by the annotators of each country against each other. We also kept track of the quality using automatic IAA metrics namely Fleiss' Kappa ($\kappa$) for Subtask~1, and Kripendorff's alpha ($\alpha$) for Subtask~2. For the first batch, we flagged all the instances of disagreement, asked the annotators to relabel them and write comments in case these flagged instances were deemed as not valid. This allowed us to have a better assessment of the reasons for disagreement, and provide the annotators with tailored feedback accordingly.

For the following three batches, we tried to categorize clear patterns of disagreement between the annotators (e.g., an annotator systematically disagreeing with the other annotators) and discussed them individually with the annotators to rectify them in future batches. We have only asked them to relabel the samples of high disagreement in case we could not determine a pattern for the disagreement.
For the last batch, we resorted to asking the annotators to relabel the samples of disagreement, to get an approximate evaluation of the impact of this process on the aggregated labels. 

\paragraph{Analysis of the IAA Scores} Table~\ref{tab:IAA_detailed} demonstrates how the IAA scores (Fleiss' Kappa for Subtask~1, and Krippendorff's Alpha for Subtask~2) changed as the annotation process progressed. First, the values hint at acceptable levels of agreement between the annotators for both subtasks. However, we notice that the range of the IAA scores differs from one country to another, especially for Subtask~1. The variation in the ranges of the IAA scores could be attributed to (a) the level of homogeneity between the dialects spoken in each country, and (b) the annotators' representativeness/knowledge of the different dialects spoken in their countries. Recruiting annotators from different regions within the same country (e.g., the case of the Algerian annotators), could increase the possibility of disagreement compared to when they all came from the same region (e.g., the case of the Egyptian annotators, where all are from Cairo).

Regarding the annotators' performance, we notice that the agreement between the annotators categorically increased by asking them to reannotate the high-disagreement sentences for their validity in their country-level dialects (Subtask~1). That said, the impact of this relabeling process on the number of valid sentences according to the majority voting is minimal for the last annotation batches. This increase in the agreement scores post-relabeling was not as consistent for the ALDi levels (Subtask~2), in which we sometimes notice insignificant decreases. This could be attributed to the subjectivity of the ALDi Estimation task, compared to the Validity task. Lastly, the annotators' performance, measured by the IAA scores, was consistent across the different annotation batches, showcasing the effectiveness of our process.




\section{Detailed Analysis of Subtask 1 Results}
\label{sec:detailed_subtask1_results}
\begin{table*}[tb]

\begin{subtable}[c]{0.58\textwidth}
    \scriptsize
    \centering
    \begin{tabular}{clcccc}
    \toprule
        \multirow{2}{*}{\textbf{Rank}}& \multirow{2}{*}{\textbf{System}} & \multicolumn{4}{c}{\textbf{Macro-average}}\\ \cmidrule{3-6}
        & & \textbf{Acc. ($\uparrow$)}& \textbf{Prec. ($\uparrow$)}& \textbf{Recall ($\uparrow$)}& \textbf{\texttt{F\textsubscript{1}} score ($\uparrow$)}\\
        \midrule
            1 & \multirow{1}{*}{\textbf{Elyadata}} & 69.27\textsubscript{±4.3} & 43.07\textsubscript{±11.0} & \textbf{62.17\textsubscript{±5.6}} & \textbf{49.85\textsubscript{±8.3}}\\

            \textcolor{blue}{BL I} & \multirow{1}{*}{Top 90\%} & 75.99\textsubscript{±6.2} & 57.08\textsubscript{±15.4} & 41.92\textsubscript{±14.9} & 45.21\textsubscript{±10.3}\\

            2& \multirow{1}{*}{\textbf{NLP\_DI}} & 74.41\textsubscript{±6.2} & 49.56\textsubscript{±11.0} & 39.70\textsubscript{±11.4} & 43.02\textsubscript{±9.3}\\

            \textcolor{blue}{BL II}& \multirow{1}{*}{Random} & 50.06\textsubscript{±1.8} & 26.09\textsubscript{±9.3} & 49.92\textsubscript{±3.3} & 33.31\textsubscript{±8.4}\\

            \textcolor{blue}{BL III} & \multirow{1}{*}{Top 1} & \textbf{77.40\textsubscript{±8.0}} & \textbf{75.20\textsubscript{±11.3}} & 20.52\textsubscript{±11.7} & 30.37\textsubscript{±12.8}\\

            3 & \multirow{1}{*}{\textbf{dzNlp}} & 75.42\textsubscript{±7.7} & 61.21\textsubscript{±11.7} & 15.17\textsubscript{±5.3} & 23.61\textsubscript{±6.7}\\

        \bottomrule
    \end{tabular}%
    \subcaption{DA samples.}
\end{subtable}
~
\begin{subtable}[c]{0.35\textwidth}
    \scriptsize
    \centering

    \begin{tabular}{cccc}
        \toprule
        \multicolumn{4}{c}{\textbf{Macro-average}}\\ \cmidrule{1-4}
        \textbf{Acc. ($\uparrow$)}& \textbf{Prec. ($\uparrow$)}& \textbf{Recall ($\uparrow$)}& \textbf{\texttt{F\textsubscript{1}} score ($\uparrow$)}\\
        \midrule
        40.68\textsubscript{±9.4} & 95.92\textsubscript{±3.4} & 39.02\textsubscript{±10.6} & 54.58\textsubscript{±9.1}\\
        34.23\textsubscript{±15.3} & 96.43\textsubscript{±4.2} & 30.75\textsubscript{±16.2} & 44.65\textsubscript{±17.8}\\
        33.51\textsubscript{±15.2} & 97.40\textsubscript{±3.1} & 30.56\textsubscript{±15.6} & 44.48\textsubscript{±16.1}\\
        \textbf{51.43\textsubscript{±5.6}} & 94.36\textsubscript{±4.5} & \textbf{51.34\textsubscript{±6.0}} & \textbf{66.30\textsubscript{±5.5}}\\
        13.26\textsubscript{±7.7} & \textbf{100.00\textsubscript{±0.0}} & 7.81\textsubscript{±7.4} & 13.71\textsubscript{±11.3}\\
        10.22\textsubscript{±5.3} & 85.19\textsubscript{±31.9} & 5.22\textsubscript{±4.4} & 9.49\textsubscript{±7.7}\\
        \bottomrule
    \end{tabular}%
    \subcaption{MSA samples.}
\end{subtable}

\caption{The performance of the systems submitted to Subtask 1 on the DA and MSA samples of the test set. The systems are ordered according to their macro-averaged F1 scores on the whole test set as indicated in Table~\ref{tab:subtask1_f1}.}
\label{tab:subtask1_perf_msa_da}
\end{table*}

\begin{table*}[tb]

\begin{subtable}[c]{0.58\textwidth}
    \scriptsize
    \centering
    \begin{tabular}{clcccc}
    \toprule
        \multirow{2}{*}{\textbf{Rank}}& \multirow{2}{*}{\textbf{System}} & \multicolumn{4}{c}{\textbf{Macro-average}} \\ \cmidrule{3-6} 
        & & \textbf{Acc. ($\uparrow$)}& \textbf{Prec. ($\uparrow$)}& \textbf{Recall ($\uparrow$)}& \textbf{\texttt{F\textsubscript{1}} score ($\uparrow$)}\\
        \midrule
            1 & \multirow{1}{*}{\textbf{Elyadata}} & 68.02\textsubscript{±4.1} & 52.25\textsubscript{±12.0} & 67.16\textsubscript{±5.4} & \textbf{58.18\textsubscript{±8.9}}\\

            \textcolor{blue}{BL I} & \multirow{1}{*}{Top 90\%} & \textbf{73.07\textsubscript{±4.7}} & 62.54\textsubscript{±14.0} & 54.28\textsubscript{±14.8} & 56.16\textsubscript{±12.3}\\

            2& \multirow{1}{*}{\textbf{NLP\_DI}} & 71.73\textsubscript{±4.6} & 57.50\textsubscript{±12.2} & 49.65\textsubscript{±13.3} & 53.09\textsubscript{±12.7}\\

            \textcolor{blue}{BL II}& \multirow{1}{*}{Random} & 46.91\textsubscript{±5.8} & 35.52\textsubscript{±9.8} & \textbf{69.01\textsubscript{±15.3}} & 46.19\textsubscript{±10.9}\\

            \textcolor{blue}{BL III} & \multirow{1}{*}{Top 1} & 72.52\textsubscript{±7.5} & \textbf{77.74\textsubscript{±13.6}} & 28.87\textsubscript{±12.5} & 40.25\textsubscript{±14.8}\\

            3 & \multirow{1}{*}{\textbf{dzNlp}} & 69.94\textsubscript{±7.6} & 68.39\textsubscript{±10.0} & 21.85\textsubscript{±9.1} & 32.42\textsubscript{±11.5}\\

        \bottomrule
    \end{tabular}%
\end{subtable}
~
\begin{subtable}[c]{0.35\textwidth}
    \scriptsize
    \centering

    \begin{tabular}{ccccc}
        \toprule
        \multicolumn{5}{c}{\textbf{Individual Region \texttt{F\textsubscript{1}} score ($\uparrow$)}}\\ \cmidrule{1-5}
        \textbf{Maghreb\textsubscript{3}} &  \textbf{Nile\textsubscript{2}} &  \textbf{Levant\textsubscript{2}} &  \textbf{Gulf\textsubscript{1}} &  \textbf{Gulf of Aden\textsubscript{1}} \\
        
        \midrule
        55.42 &       68.54 &   \textbf{67.81} & 45.21 &      \textbf{53.89} \\
        \textbf{61.08} &       \textbf{69.58} &   64.20 & \textbf{51.17} &      34.76 \\
        54.71 &       67.97 &   65.65 & 36.69 &      40.43 \\
        45.59 &       56.40 &   58.06 & 27.91 &      43.00 \\
        50.59 &       60.45 &   31.84 & 40.55 &      17.80 \\
        44.14 &       44.27 &   34.09 & 25.55 &      14.03 \\
        \bottomrule
    \end{tabular}%

\end{subtable}

\caption{The performance of the systems submitted to Subtask 1, in predicting multi-label macro-regional dialects for the DA samples of the test set. In addition to the Macro-average F1 score, the individual f1 score for each region is reported. \textbf{Note:} the countries representing the regions are: \textit{Maghreb} (Algeria, Tunisia, Morocco), \textit{Nile} (Egypt, Sudan), \textit{Levant} (Palestine, Syria), \textit{Gulf} (Iraq), and \textit{Gulf of Aden} (Yemen).}
\label{tab:subtask1_regional_f1}
\end{table*}

As described in §\ref{sec:sampling_sentences}, 75 out of the 1,120 samples used to form the development and test sets for Subtasks 1 and 2 are automatically identified as being in MSA. For Subtask 1, these MSA samples are expected to be labeled as valid in all the considered dialects. On checking the validity labels for these samples, we indeed found that they are mostly deemed valid in all the considered country-level dialects. The developed systems are expected to accordingly predict that these sentences are valid in all the considered dialects.

Consequently, we report their performance on the automatically identified DA, and MSA samples respectively in Table~\ref{tab:subtask1_perf_msa_da}. Since the MSA samples represent a small proportion of the development and test sets, we find that the models' performance on the DA samples is not different from their overall performance reported in Table~\ref{tab:subtask1_f1}.

For the MSA samples, we notice that the macro-average Recall needs to be improved. A two-stage solution could be proposed in which a classifier first identifies if a sentence is in MSA or DA. MSA sentences can be predicted to be valid in all the considered dialects with high accuracy. Conversely, the validity labels for the DA samples could be identified using another multilabel dialect identification system.

\paragraph{Regional Level Performance} The results in Tables~\ref{tab:subtask1_f1},~\ref{tab:subtask1_perf_msa_da} indicate that there is room for improvement for the multi-label ADI systems to be reliably able to accurately operate on the country-level. Consequently, we group the nine country labels of the test set into macro-regional dialects according to \cite{baimukan-etal-2022-hierarchical} as follows: \textit{Maghreb} (Algeria, Tunisia, Morocco), \textit{Nile Basin} (Egypt, Sudan), \textit{Levant} (Palestine, Syria), \textit{Gulf} (Iraq), and \textit{Gulf of Aden} (Yemen). For each region, a sample is considered valid in the region if it is valid in any of the region's countries for which we have validity labels. For example, a sample annotated as valid in Algeria, Tunisia, and Sudan will be considered valid in \textit{Maghreb} and \textit{Nile Basin}. We similarly consider the systems' predictions for the same nine countries and aggregate them into macro-regional dialects.

The models' performance predicting the macro-regional dialects is higher than that for country-level ones as per Table~\ref{tab:subtask1_regional_f1}. However, the improvement is not as high as might have been expected, indicating that even multi-label macro-regional dialect identification is a challenging task. In the future, we plan to extend the labels in our test set to cover more countries, especially from the \textit{Gulf} region.

\end{document}